\documentclass[final,3p,times,numbers]{elsarticle}

\usepackage{amssymb}
\usepackage{amsmath, xparse}
\usepackage[ruled,vlined,linesnumbered]{algorithm2e}
\SetKw{Continue}{continue}
\usepackage{graphicx}
\usepackage{multirow}
\usepackage{stfloats}
\usepackage{amsmath,amsfonts}
\usepackage{array}
\usepackage[caption=false,font=normalsize,labelfont=sf,textfont=sf]{subfig}
\usepackage{textcomp}
\usepackage{url}
\usepackage{verbatim}
\usepackage{threeparttable}
\usepackage{color}
\usepackage{mathrsfs}
\usepackage{makecell}
\usepackage{tabularx}
\usepackage{booktabs}
\usepackage{bbding}
\usepackage[table]{xcolor}

\usepackage[backref=true]{hyperref}

\usepackage{tikz,xcolor,hyperref}

\definecolor{lime}{HTML}{A6CE39}
\DeclareRobustCommand{\orcidicon}{
\begin{tikzpicture}
\draw[lime, fill=lime] (0,0)
circle[radius=0.16]
node[white]{{\fontfamily{qag}\selectfont \tiny \.{I}D}}; 
\end{tikzpicture}
\hspace{-2mm}
}
\foreach \x in {A, ..., Z}{%
\expandafter\xdef\csname orcid\x\endcsname{\noexpand\href{https://orcid.org/\csname orcidauthor\x\endcsname}{\noexpand\orcidicon}}
}

\journal{Expert Systems With Applications}

\begin{document}

\begin{frontmatter}



\title{
AdvReal: Physical Adversarial Patch Generation Framework for Security Evaluation of Object Detection Systems
}


\author[label1,label2]{Yuanhao Huang}
\ead{yuanhao\_huang@buaa.edu.cn}

\author[label1,label3]{Yilong Ren\corref{cor1}}
\ead{yilongren@buaa.edu.cn}

\author[label2,label4]{Jinlei Wang}
\ead{20231800822@imut.edu.cn}

\author[label2,label4]{Lujia Huo}
\ead{20231100544@imut.edu.cn}

\author[label1,label2]{Xuesong Bai}
\ead{xs\_bai@buaa.edu.cn}

\author[label1]{Jinchuan Zhang}
\ead{22376270@buaa.edu.cn}

\author[label1,label3]{Haiyang Yu}
\ead{hyyu@buaa.edu.cn}


\cortext[cor1]{Corresponding author}

\affiliation[label1]{organization={School of Transportation Science and Engineering, Beihang University},
            addressline={Kejiyuan Rd}, 
            city={Haidian District},
            postcode={100191}, 
            state={Beijing},
            country={P.R China}}

\affiliation[label2]{organization={State Key Lab of Intelligent Transportation System},
            addressline={Kejiyuan Rd}, 
            city={Haidian District},
            postcode={100191}, 
            state={Beijing},
            country={P.R China}}
            
\affiliation[label3]{organization={Zhongguancun Laboratory},
            city={Haidian District},
            postcode={100191}, 
            state={Beijing},
            country={P.R China}}
            
\affiliation[label4]{organization={Aviation Academy, Inner Mongolia University of Technology},
            addressline={Aimin Street}, 
            city={Hohhot},
            postcode={010051}, 
            state={Inner Mongolia},
            country={P.R China}}

\begin{abstract}

Autonomous vehicles are typical complex intelligent systems with artificial intelligence at their core. However, perception methods based on deep learning are extremely vulnerable to adversarial samples, resulting in security accidents. How to generate effective adversarial examples in the physical world and evaluate object detection systems is a huge challenge. In this study, we propose a unified joint adversarial training framework for both 2D and 3D domains, which simultaneously optimizes texture maps in 2D image and 3D mesh spaces to better address intra-class diversity and real-world environmental variations. The framework includes a novel realistic enhanced adversarial module, with time-space and relighting mapping pipeline that adjusts illumination consistency between adversarial patches and target garments under varied viewpoints. Building upon this, we develop a realism enhancement mechanism that incorporates non-rigid deformation modeling and texture remapping to ensure alignment with the human body’s non-rigid surfaces in 3D scenes. Extensive experiment results in digital and physical environments demonstrate that the adversarial textures generated by our method can effectively mislead the target detection model. Specifically, our method achieves an average attack success rate (ASR) of 70.13\% on YOLOv12 in physical scenarios, significantly outperforming existing methods such as T-SEA (21.65\%) and AdvTexture (19.70\%). Moreover, the proposed method maintains stable ASR across multiple viewpoints and distances, with an average attack success rate exceeding 90\% under both frontal and oblique views at a distance of 4 meters. This confirms the method’s strong robustness and transferability under multi-angle attacks, varying lighting conditions, and real-world distances. The demo video and code can be obtained at ~\url{https://github.com/Huangyh98/AdvReal.git}.
\end{abstract}



\end{frontmatter}



\section{Introduction}

With the rapid advancement of artificial intelligence (AI), autonomous vehicles (AVs) have become widely adopted worldwide. Drivable area detection and object detection are the primary components in any autonomous driving (AD) technology~\cite{zhang2022intelligent,giri2025so}. Although these algorithms have achieved significant breakthroughs in performance, the frequent traffic accidents involving autonomous vehicles have raised public concerns about the security and robustness of their perception systems. Consequently, the systematic testing and evaluation of object detection algorithms have become a prominent research focus.

Currently, AD object detection systems face two primary risks: natural scenario risks and adversarial scenario risks. Natural scenario risks stem from variations in the real environment, such as changes in lighting, weather, and the behavior of traffic participants~\cite{swerdlow2024street}. The performance of perception systems is typically simulated and assessed through scenario generation techniques. In contrast, adversarial scenario risks involve malicious attackers employing adversarial attacks to disrupt the normal operation of perception systems in both digital and physical domains~\cite{huang2025advswap, ran2025black, hu2023physically}. Notably, physical attacks directly target objects in the real world, effectively bridging the digital and physical realms and resulting in more impactful consequences. Such attacks can cause the autonomous driving system to fail in accurately detecting and localizing objects, which in turn may lead to hazardous driving decisions~\cite{fang2024state, bai2024ar}.

Physical adversarial patches can be used to effectively evaluate the security of object detection systems~\cite{zhang2021evaluating}. The adversarial patch for pedestrian detection was first introduced by AdvPatch~\cite{brown2017adversarial}. By optimizing and generating adversarial patches, this approach effectively evades target detectors in the physical world, although it is limited to static objects. Subsequently, researchers have proposed a variety of patch-based physical adversarial attack methods, such as AdvTshirt~\cite{xu2020adversarial}, AdvTexture~\cite{hu2022adversarial}, and T-SEA~\cite{huang2023t}. In addition, techniques like NatPatch~\cite{hu2021naturalistic} and AdvCaT~\cite{hu2023physically} have been developed for natural-looking and camouflaged attacks. While previous studies demonstrated strong attack performance in constrained digital settings, they often neglected the complex environmental factors encountered in real-world applications.

Researchers have found that the threat posed by patch-based attacks may be smaller than previously believed, and that the success rate observed in simple digital simulations does not necessarily translate to real-world effectiveness~\cite{hingun2023reap}. In pedestrian detection tasks, factors such as intra-class diversity, ambient lighting, and clothing wrinkles can all diminish the performance of adversarial patches~\cite{wang2024attention}. For instance, applying patches to three-dimensional objects may lead to geometric distortions due to perspective projection effects~\cite{mahima2024toward}. The discrepancy between simplified adversarial training in digital environments and the complex, dynamic nature of the physical world is one of the primary challenges hindering the real-world deployment of adversarial patches~\cite{cui2024adversarial}.

\begin{figure}[t!]
\centering
\includegraphics[width=10cm]{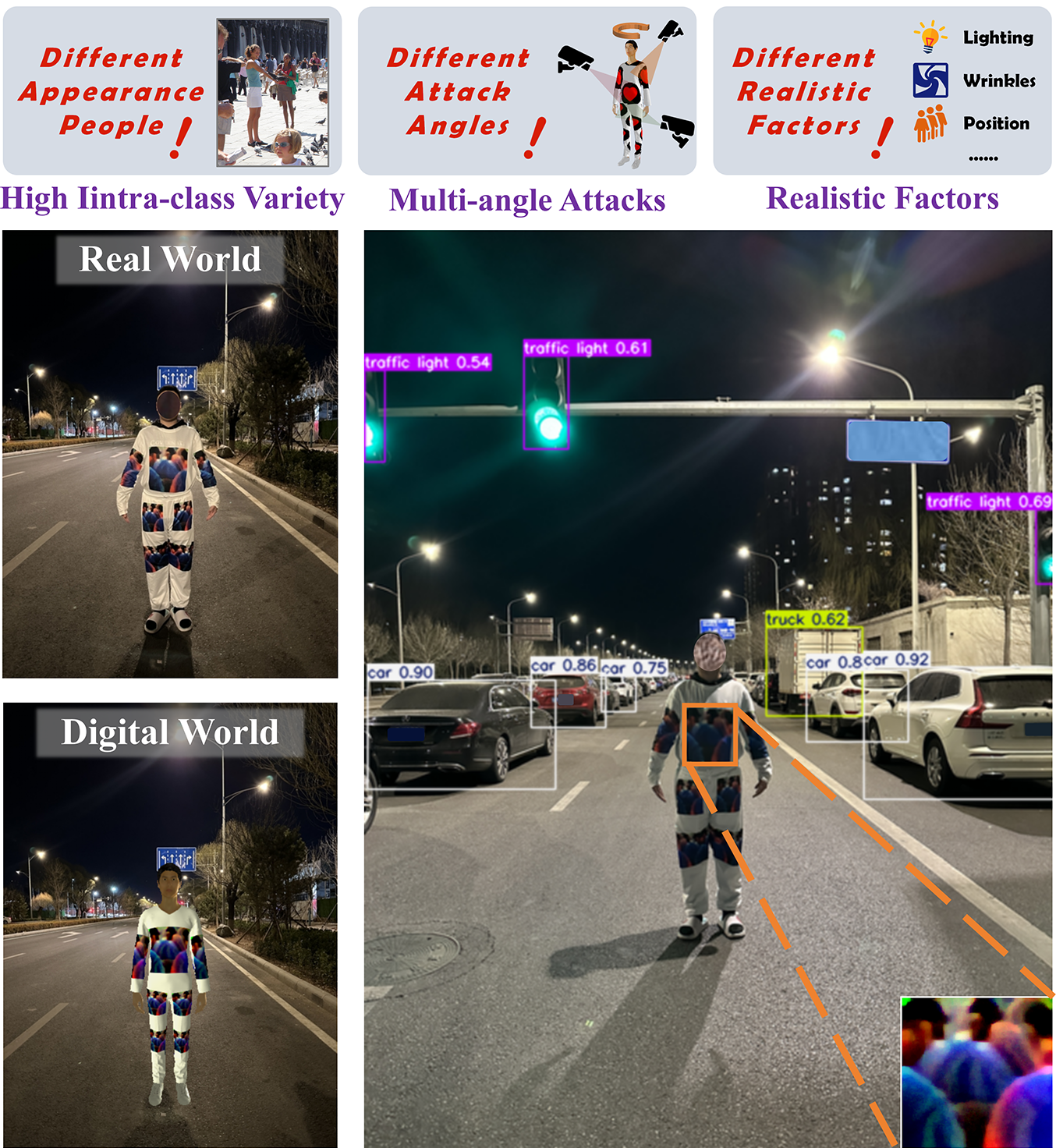}
\caption{In response to multi-dimensional realistic challenges, performance of proposed adversarial patches in the physical world.}
\label{cover}
\vspace{-4 mm} 
\end{figure}

In this study, we proposed a reality-enhanced adversarial training framework called AdvReal. Through realistic enhancement and joint optimization of samples in both two-dimensional (2D) and three-dimensional (3D) spaces, more effective adversarial patches are generated for the physical world. The framework comprises three modules. The patch adversarial module refines patches in 2D and synthesizes samples that capture intra-class diversity. The realistic enhancement module processes 3D models to produce lifelike adversarial samples. And the joint optimization module concurrently refines both 2D and 3D adversarial samples to yield robust patches for complex physical world scenarios, shown in Fig.~\ref{cover}. Our goal is to generate adversarial patches to evaluate and expose vulnerabilities in perception systems deployed in AVs, especially pedestrian detection modules, under real-world environmental conditions. The main contributions of this work are as follows:

\begin{itemize}
\item [1)] Designed a general joint adversarial training framework for both 2D and 3D samples to overcome the challenges posed by intra-class diversity and variations in real-world environments during patch training. The framework supports the simulation of realistic adversarial threats against AV perception models by capturing differences in viewpoint, distance, and body appearance common in urban driving scenarios.

\item[2)] Proposed an adversarial sample reality enhancement method based on non-rigid surfaces modeling and a realistic matching mechanism in 3D space. By incorporating realistic elements such as cloth deformation and varying illumination, the method better reflects how adversarial patches would appear on pedestrians in actual AV perception settings.

\item [3)] Numerous sets of experiments were conducted in both digital and physical environments, and we analyzed the underlying attack mechanism of the patch using visualization methods. Our quantitative and qualitative results show that the generated adversarial patch achieves a higher attack success rate, improved robustness, and superior transferability against pedestrian detectors compared to existing methods. The results offer actionable insights for improving detector architectures in AV systems and contribute to the broader goal of AI security evaluation in autonomous driving.
\end{itemize}

The remainder of this paper is organized as follows. Section 2 surveys related work on adversarial patches, the evasion of person detectors, and the practical challenges posed by adversarial attacks. Section 3 formalizes the research problem and presents the proposed methodology. Section 4 describes the simulation environment and experimental setup, and reports results obtained in both digital and physical settings. Section 5 discusses the advantages and limitations of the approach in light of the experimental findings. Finally, Section 6 concludes the study and outlines directions for future research.

\section{Related works}

\subsection{Adversarial Patch}



Understanding the behavior of adversarial attackers has become a growing focus in AI security research. Recent works have proposed systematic taxonomies to describe attacker characteristics, including access level (e.g., glass-box, gray-box, closed-box), behavior characteristics (e.g., digital and physical attack), attack intention (e.g., evasion, impersonation, sabotage, et al.), and attack capability (e.g., attack success rate, transferability, robustness, et al.)~\cite{nguyen2025survey}. In adversarial attack tasks, a \textit{glass-box} (or white-box) setting assumes the attacker has full access to the victim model's parameters and gradients, whereas a \textit{closed-box} (or black-box) setting assumes no internal knowledge of the model is available, and the attacker can only interact with the detector via input-output queries. The frameworks help researchers to analyze attack feasibility, threat severity, and defense applicability across different deployment scenarios. In safety-critical systems such as autonomous driving, distinguishing attacker intent and capability is especially important, as it directly impacts how risks should be assessed and mitigated~\cite{wang2023adversarial}. The background provides a foundation for studying concrete physical-world attacks such as adversarial patches.



Compared to digital perturbations added directly to images~\cite{xiao2025transformer}, adversarial patches present a unique capability to attack detectors in the physical world. Since these patches are installed before the camera imaging process, they pose a greater threat to autonomous driving perception systems and are of greater practical significance.

As a typical sparse perturbation, an adversarial patch can effectively degrade classification performance across various inputs~\cite{williams2024camopatch}. In the context of traffic sign detection, Zhou et al. investigated how adversarial patches can manipulate AVs by making traffic signs unrecognizable or misclassified, showcasing the real-world implications of such attacks~\cite{zhou2024stealthy}. Adversarial patches are also widely used in face recognition, where wearable masks~\cite{liu2024eap} or even tiny stickers~\cite{wei2022simultaneously} can be generated to mislead deep neural networks. Therefore, adversarial patches have been widely used to evaluate the security and reliability of object detection systems.

\subsection{Evade Person Detectors with Adversarial Patch}



Humans are among the most critical traffic participants in autonomous driving environments. Using adversarial patches to attack pedestrian perception modules in AV systems provides an effective means to evaluate their susceptibility to adversarial risks. Unlike objects such as traffic signs, which exhibit minimal intra-class variability, humans present significant intra-class diversity, making the detection task more challenging~\cite{abed2024deep} and increasing the difficulty of ensuring robust and reliable performance under adversarial conditions.

A common approach involves optimizing adversarial patches by placing them on the human body and minimizing the maximum bounding-box probability associated with pedestrian detection. AdvPatch was the first adversarial patch method designed to fool pedestrian detectors by directly optimizing detection loss~\cite{thys2019fooling}. Similarly, TOG employs attack-specific gradients derived from adversarial targets to cause detectors to either miss or misclassify pedestrians~\cite{chow2020adversarial}. However, TOG does not account for deployment in physical-world settings. Furthermore, a GAN-based generative strategy has been introduced to constrain the appearance of generated patches, ensuring they remain natural and inconspicuous~\cite{guesmi2024dap}. NatPatch follows a similar principle by sampling optimal images from GANs~\cite{hu2021naturalistic}, resulting in patches that are both visually realistic and adversarially effective. While the above methods offer promising results in static image-based evaluations, they fall short of addressing the full complexity of AV perception systems, which must operate under dynamic viewpoints, environmental variations, and real-time physical constraints. As such, their ability to evaluate adversarial robustness in practical autonomous driving scenarios remains limited.

\subsection{Multiple realistic challenges}

When adversarial patches are applied in the physical world, their attack effectiveness is significantly diminished or rendered entirely ineffective. This limitation arises because most prior studies have overlooked three critical real-world factors.

\subsubsection{\textbf{High level of intra-class variety}} 
Pedestrians exhibit a high degree of intra-class variability, characterized by significant differences in appearance, posture, and dynamic motion. In autonomous driving perception systems, the variability poses major challenges for consistent and reliable pedestrian detection under real-world conditions. To address these variations, the SOTA pedestrian detection models leverage extensive training datasets and advanced feature extraction techniques to improve algorithmic robustness~\cite{ren2016faster, redmon2018yolov3, carion2020end}. Consequently, devising effective and generalized adversarial attack strategies requires exceptional adaptability and resilience. Simen Thys et al. were among the first to explore such attacks on targets with high intra-class variability, utilizing real images of diverse individuals~\cite{thys2019fooling}. Building on this, Zhou et al. proposed the PosePatch framework, a patch adaptation network for adversarial patch synthesis guided by perspective transformations and estimated human poses~\cite{zhou2025fooling}. The efforts reveal the necessity of accounting for real-world appearance diversity when evaluating the AI security of pedestrian detection modules in AVs.

\subsubsection{\textbf{Multi-angle attacks}} 
Traditional superposition training ensures that pedestrian detectors perform effectively when facing the patch but lacks robustness to varied angles or occlusions. In the context of AVs, where onboard cameras observe pedestrians from dynamic and shifting perspectives, adversarial patches must be effective across a wide range of viewing angles to pose a realistic threat. To address these challenges, AdvTexture introduces a scalable adversarial texture generation method based on ring cropping~\cite{hu2022adversarial}. This approach can be applied to clothing of arbitrary shapes, allowing individuals wearing these garments to evade detection by human detectors from various angles. Furthermore, Arman Maesumi et al. proposed a framework for adversarial training using 3D modeling, which optimizes 3D adversarial cloaks through variations in poses, spatial positions, and camera angles~\cite{maesumi2021learning}. By conducting multi-angle adversarial training, the effectiveness of patches in multi-detection angle scenarios can be significantly improved~\cite{li2025uv, hu2023physically}. Such techniques are critical for testing the vulnerability of real-world pedestrian detection systems deployed in autonomous vehicles.

\subsubsection{\textbf{Realistic factors}} 
Accurately capturing all real-world variations through digital images of simulated patches for adversarial training and evaluation remains challenging~\cite{hingun2023reap}. Modeling non-rigid object surfaces, such as clothing, during training and utilizing them for adversarial optimization poses a considerable challenge. For instance, Adv-Tshirt employs video frames of a person wearing a T-shirt with a checkerboard pattern as training data, mapping the patch onto a distorted checkerboard for training~\cite{xu2020adversarial}. AdvCaT proposes a non-rigid deformation method based on TopoProj to simulate wrinkles on human clothing~\cite{hu2023physically}. Integrating such non-rigid modeling into the patch optimization process significantly improves robustness under real-world physical constraints.  Traditional patch-based methods often assume that patches are square, axis-aligned, and freely placed within the image. However, random rotation, cropping, and additive noise do not fully simulate complex environmental variations such as illumination shifts, pose changes, and background clutter. REAP proposes an improved method that incorporates geometric and lighting transformations to enable large-scale performance evaluation of patches, particularly traffic signal signs~\cite{hingun2023reap}. To ensure safety and reliability in autonomous driving applications, it is essential to include physically realistic transformations in patch training pipelines to evaluate adversarial robustness under dynamic AV scenarios.

\section{Methodology}

In this section, we first formulate the object and define the optimization problem. Then, we demonstrate the details of the proposed method, including adversarial generation framework, realistic matching module, and joint optimization strategy.

\begin{figure*}[t!]
\centering
\includegraphics[width=12cm]{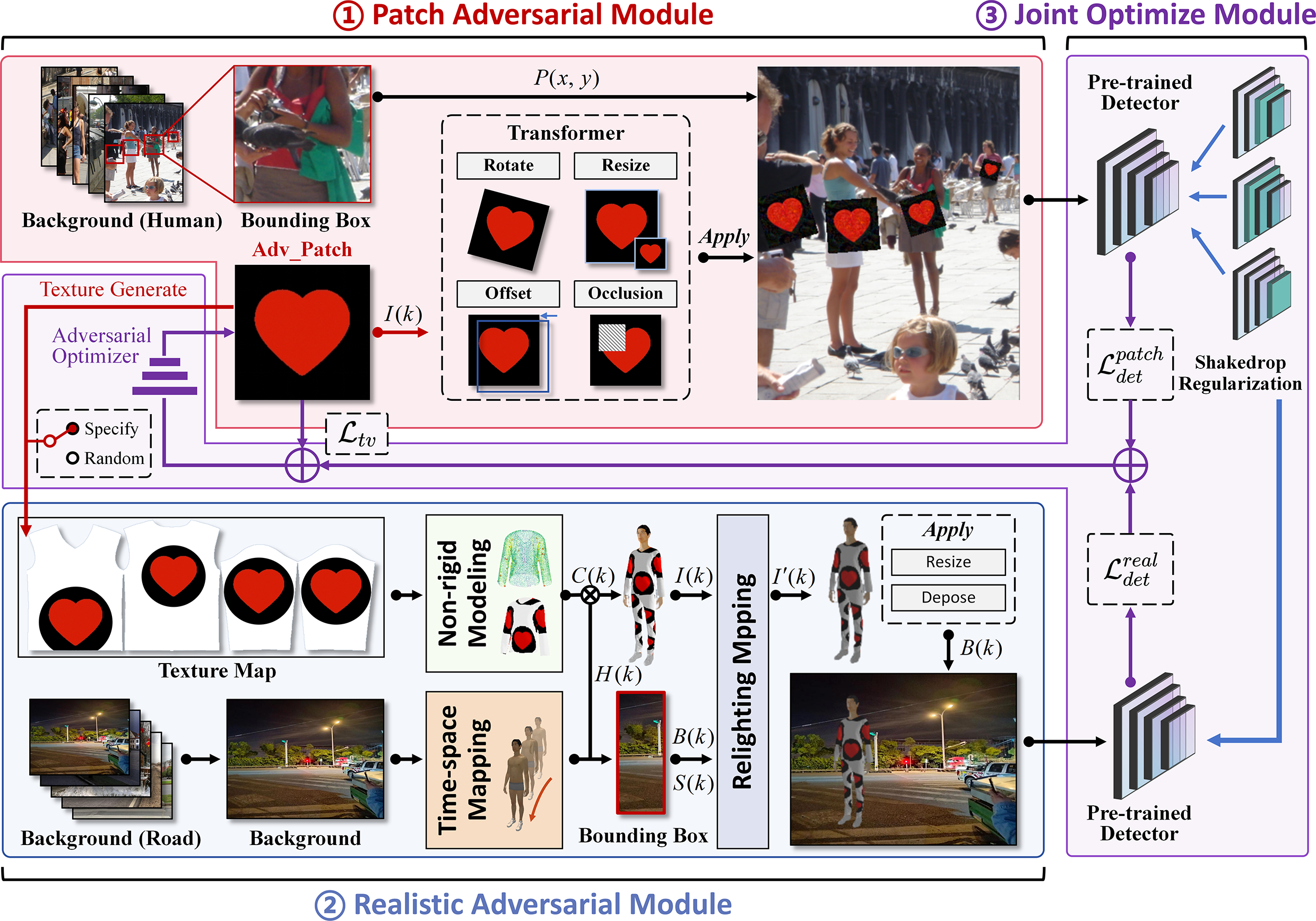}
\caption{Overview of the proposed adversarial attack method.}
\label{02_framework}
\vspace{-4 mm} 
\end{figure*}

\subsection{Problem Definition}

In this study, we perform a joint adversarial attack on both 2D high-level intra-class variability data and 3D augmented reality data. The ultimate objective is to effectively reduce the detection accuracy of adversarial patches recognized by human detectors in real-world settings. Specifically, let \( P(X, Y) \) denote the joint distribution of pedestrian data, where \( X \subseteq \mathbb{R}^d \) is the input image space and \( Y \subseteq \{1, \ldots, C\} \) is the label space. A pretrained pedestrian detector \( f_h: [-1, 1]^d \to [0, 1]^C \) maps an input \( x \in X \) to a probability distribution over \( C \) classes, with the predicted label \( \hat{y} = \arg\max_{c} f_c(x) \). Given a sample set \( \{(x_1, y_1), \ldots, (x_n, y_n)\} \sim P(X, Y) \), the goal of the adversarial attack is to construct a perturbed input \( x^{\text{adv}} = x + \delta \), where \( \|\delta\|_p \leq \epsilon \), such that the detector \( f(x^{\text{adv}}) \) misclassifies \( x^{\text{adv}} \) into a label \( \hat{y} \neq y \).

The corresponding optimization problem is formulated in Eq.~\ref{eq1}:

\begin{equation}
\delta^* = \underset{\delta}{\arg\min} \; \frac{1}{n} \sum_{i=1}^{n} \mathcal{L}(f_h(x_i + \delta), y_i) + \lambda \mathcal{R}(\delta),
\label{eq1}
\end{equation}

where \( \mathcal{L}(f(x_i + \delta), y_i) \) measures the discrepancy between the detector's prediction on the perturbed input \( x_i + \delta \) and the true label \( y_i \). The regularization term \( \mathcal{R}(\delta) \) enforces the smoothness of the adversarial patch, thereby reducing visual artifacts and improving both its generalization and transferability by preventing overfitting. The coefficient \( \lambda \) balances the adversarial loss and the regularization term, ensuring that the adversarial patch is not only effective in misleading the detector but also robust across diverse real-world scenarios and detection architectures.

\subsection{Adversarial Generation Framework}

To address the limitations of existing methods, this paper introduces a novel framework for generating adversarial patches designed to deceive human detectors in real-world scenarios. The proposed framework, named AdvReal, performs joint optimization over 2D and 3D adversarial losses, effectively tackling challenges such as high intra-class variability, multi-angle attack scenarios, and complex real-world factors. Unlike traditional adversarial patches applied to traffic signs or rigid materials, our method focuses on patches applied to non-rigid clothing surfaces, ensuring effectiveness from multiple angles in realistic scenarios.

As illustrated in Fig.~\ref{02_framework}, the AdvReal framework consists of three key modules: the Patch Adversarial Module, the Realistic Enhancement Module, and the Joint Optimization Module. These components function in synergy to improve both the adversarial strength and the physical plausibility of the generated patches.

\subsubsection{Patch Adversarial Module} 
To enhance the generalization performance of intra-class adversarial attacks against human detectors, the Patch Adversarial Module leverages a diverse human dataset for 2D adversarial training. In this module, patches are placed at the center of the human bounding boxes within the dataset, followed by transformation operations such as rotation, resizing, offsetting, and occlusion. These transformations are applied to ensure the robustness of the adversarial patches under varying conditions. The synthesized images, comprising patches overlaid on humans and their backgrounds, are then utilized for 2D adversarial training to improve attack efficacy.

\subsubsection{Realistic Enhancement Module} 
This module employs 3D meshes to model humans, clothing, and pants, attaching the texture map generated by the adversarial patch to the clothing. Using the \textbf{time-space enhancement strategy} (see Sec.~\ref{Nonrigid}), 3D adversarial garments with realistic folds are rendered. The \textbf{realistic matching mechanism} (see Sec.~\ref{Realistic}) first aligns the rendered human models to realistic spatial configurations (distance, scale, elevation, and azimuth), ensuring accurate spatial representation. Then, nonlinear adjustments to brightness and contrast are applied to simulate diverse lighting conditions, seamlessly integrating geometric and illumination variations to produce highly realistic adversarial images. Finally, the resized human images are synthesized onto decomposed background images for 3D adversarial training. Importantly, the Patch Adversarial Module and the Realistic Enhancement Module operate synchronously within each training batch.

\subsubsection{Joint Optimization Module} 
The synthetic images produced by the Patch Adversarial Module and the Realistic Enhancement Module are fed into the pre-trained human detector targeted for attack. During this process, certain layers of the detector are randomly frozen, and the output is used to compute the adversarial loss. The appearance of the adversarial patch is iteratively optimized by minimizing a combined loss function, which includes 2D detection loss, 3D detection loss, and Total Variation (TV) loss. The \textbf{joint optimization mechanism} (see Sec.~\ref{Joint}) ensures that the disguised patch remains effective across both 2D and 3D detection scenarios.


\subsection{Non-rigid surfaces modeling}
\label{Nonrigid}

Methods based on shear transformation or 3D thin plate spline (3D-TPS) often disregard the nonlinear deformation characteristics of real fabrics~\cite{wei2024physical}. The simplification leads to two key limitations in adversarial pattern deployment: (1) geometric discontinuities in high-curvature regions of cloth folds, and (2) physically implausible stretching artifacts during dynamic body movements, resulting in noticeable texture distortions.

To improve realism in fabric deformation, we integrate a physics-based modeling strategy that analyzes the motion flow of the clothing mesh to estimate load-bearing stress points. These stress points are used as anchors for generating natural, non-rigid folds, ensuring diverse yet realistic fabric behavior even under consistent adversarial training motions~\cite{gundogdu2020garnet}.

\begin{figure}[t!]
\centering
\includegraphics[width=8.6cm]{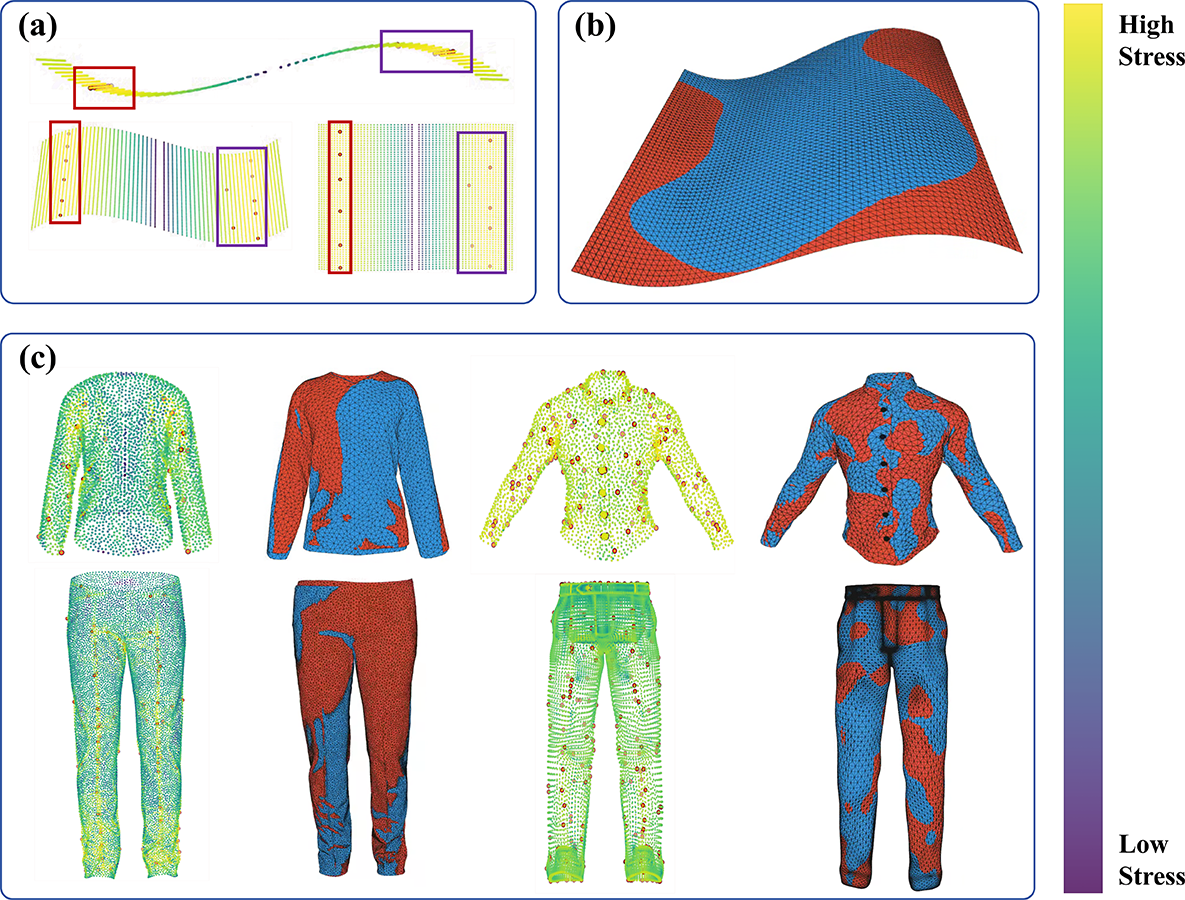}
\caption{Schematic diagram and results display of non-rigid surfaces modeling. (a) Mesh of the simulated cloth, with control points strategically distributed in high-stress regions. (b) Deformation result after applying random target points with smoothing, exhibiting no abnormal distortions. (c) Non-rigid modeling and deformation outcomes for both shirt and pants.}
\label{nonrigid}
\vspace{-4 mm} 
\end{figure}

We estimate the stress tensor at each vertex \( \mathbf{v}_i \) in the garment mesh \( \mathcal{M} = (\mathcal{V}, \mathcal{E}) \), where stress is accumulated from neighboring vertices:

\begin{equation}
\label{eq:stress}
\sigma_i = \sum_{j \in N(i)} w_{ij} \|\mathbf{v}_j - \mathbf{v}_i\|_2
\end{equation}
where \( w_{ij} \) is the adjacency weight, and \( N(i) \) denotes the neighboring vertices of \( \mathbf{v}_i \). As shown in Fig.~\ref{nonrigid}, High-stress areas correspond to fabric regions experiencing significant force transmission, typically near joints and constrained areas. Regions with significant force transmission—typically near joints and constrained areas—are identified by selecting vertices that satisfy:
\begin{equation}
\label{sigma}
\mathcal{S} = \{ p_i \mid \sigma_i > \sigma_{\text{thres}} \},
\end{equation}
where 
\begin{equation}
\begin{aligned}
\label{rank}
\quad \sigma(p_1) \geq \sigma(p_2) \geq \cdots \geq \sigma(p_M),
\end{aligned}
\end{equation}

Here, \(\sigma_{\text{thres}}\) is a predefined threshold to reduce computational complexity. After identifying the high-stress points \(\mathcal{S}\). The final control point set \(C\) is then determined by enforcing a spatial isolation criterion:

\begin{equation}
\label{C}
C = \Big\{ p_i \in S \,\Big|\, \forall\, p_j \in C,\ \|p_i - p_j\| \ge \frac{\gamma}{\sigma(p_i)} \Big\} \quad,
\end{equation}
where
\begin{equation}
\label{Cmax}
 \quad |C| = \max\Big\{N_{\max},\, \Big\lfloor \rho\,|S| \Big\rfloor\Big\}.
\end{equation}

Here, \(\gamma\) is a minimum distance factor that ensures sufficient spatial separation between control points, \(\rho\) is a density ratio that determines the target number of control points, and \(N_{\min}\) is a predetermined lower bound on the number of control points.

To generate realistic fabric wrinkles while retaining design flexibility, we inject a stochastic perturbation term into the deformation function to capture the natural variability observed under similar body movements:

\begin{equation}
\label{stochastic_tps}
f(\mathbf{v}) = \mathbf{v} + \sum_{k=1}^K \mathbf{w}_k \phi(\|\mathbf{v} - \mathbf{c}_k\|) + \delta \mathbf{r}_i
\end{equation}
where \( \mathbf{r}_i \sim \mathcal{N}(0, \Sigma) \) is a Gaussian noise vector, and \( \delta \) scales the perturbation. This ensures natural variations in generated wrinkles under similar body movements, enhancing the realism of adversarial textures.

However, unconstrained noise can produce physically implausible distortions. We regulate vertex displacements by imposing stress-dependent constraints:

\begin{equation}
\label{eq:stress_constraint}
\Delta\mathbf{v}_i = \min\left(1, \frac{\delta_{\text{max}} \cdot (1 + \lambda \sigma_i)}{\|\Delta\mathbf{v}_i\|_2}\right) \cdot (\mathbf{v}'_i - \mathbf{v}_i)
\end{equation}
where \( \lambda \) modulates the stress-dependent displacement expansion. This ensures regions with higher stress experience larger but controlled deformations, maintaining smoothness and physical consistency. 


\subsection{Realistic Matching Mechanism}
\label{Realistic}

In realistic scenarios, factors such as distance and illumination introduce challenges related to variations in scale, perspective, and lighting conditions. To enhance the realistic relevance of adversarial patch training and evaluation, we propose two complementary strategies: \textbf{time-space mapping} and \textbf{relight mapping}. First, time-space mapping is employed to generate pedestrian bounding boxes that accurately reflect real-world scales. Second, we introduce an optimization-based relight mapping approach to further enhance the visualization of the pedestrian detection model under diverse lighting conditions.

\subsubsection{\textbf{Time-space Mapping}}

\begin{figure}[ht!]
\centering
\includegraphics[width=8.6cm]{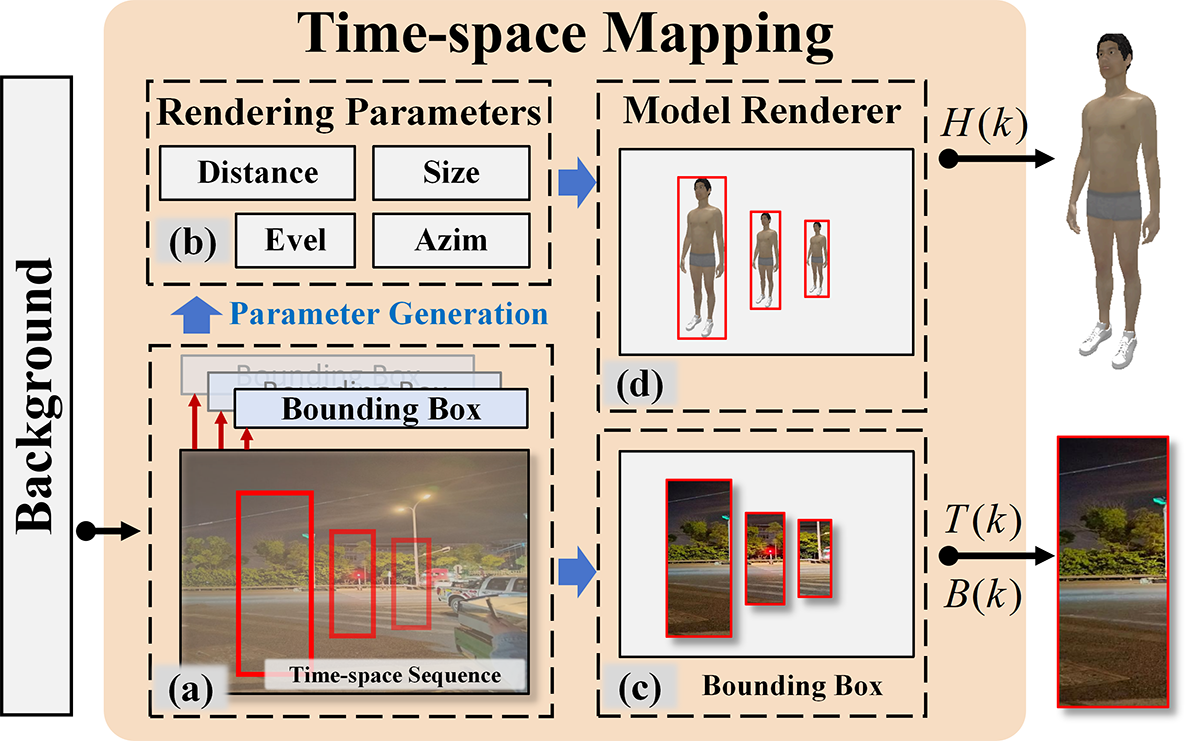}
\caption{Flowchart of spatiotemporal mapping. (a) Select consecutive candidate bounding boxes that satisfy the perspective relationship in the background image. (b) Generate rendering parameters based on the positions of the selected bounding boxes. (c) Extract background image corresponding to the bounding box. (d) Generate rendering parameters for the person within the corresponding bounding box.}
\label{time_space}
\end{figure}

We aim to render a human model that is more consistent with the time-space distribution to facilitate integration with the background.  The method first randomly select a bounding box \( b_{\text{human}} \) with human proportions. The orientation of the human model is facing, represented by a direction vector \( \mathbf{v}_{\text{orient}} \), is then determined. Based on this orientation, the bounding box is adjusted and scaled to generate two additional bounding boxes, \( b_{\text{near}} \) and \( b_{\text{far}} \), which follow a perspective-based scaling rule. Specifically, as shown in Fig.~\ref{time_space} (a), we generate a near and a far rendering image at different positions of the same background according to the position of the first bounding box and the orientation of the character model. And the near bounding box \( b_{\text{near}} \) is larger, and the far bounding box \( b_{\text{far}} \) is smaller, with their sizes inversely proportional to the distance from the camera.

As shown in Fig.~\ref{time_space} (b), the rendering parameters including scale \( s \), distance \( d \), and orientation angles \( \theta_e \) and \( \theta_a \) from the geometric properties of the boxes can be derived. The azimuth angle \( \theta_a \) is obtained by solving the inverse angle based on the facing direction of the human model.

These derived parameters guide the rendering process, where a differentiable renderer projects the adversarial texture onto a 3D human mesh, adjusting its distance, size, and azimuth. Crucially, the rendered model's silhouette (Fig.~\ref{time_space} (d)) is constrained to fit within the bounding box \( b_i \)’s coordinates (Fig.~\ref{time_space} (c)), ensuring geometric consistency when composited into the background. Subsequently, the human model $H(k)$ is used for combined rendering with the clothing $C(k)$ after non-rigid surfaces modeling. And the bounding box $B(k)$ and the background image slice $T(k)$ are used as input in Relighting mapping.

\subsubsection{\textbf{Relighting Mapping}}

In order to generate images for adversarial training, the rendered characters need to be fused with the background. The combination process can be expressed as Eq.~\ref{eq2}:

\begin{equation}
\tilde{x}^{\text{adv}}_{i+1} = M \odot \tilde{R}^{\text{hum}}_{i+1} + (1 - M) \odot X,
\label{eq2}
\end{equation}

where \( M \in \{0, 1\}^{H \times W} \) is a binary mask derived from the 3D human model, with \( M = 1 \) corresponding to the human texture and \( M = 0 \) to the background. \( M \odot \tilde{R}^{\text{hum}}_{i+1} \) overlays the rendered human texture, including the adversarial patch, onto the human region, while \( (1 - M) \odot X \) preserves the background texture.

\begin{algorithm}[t]
\KwIn{real image $I_r$, rendered image $I_0$}
\KwOut{$(\alpha^\ast,\beta^\ast,\theta^\ast)$}

\textbf{Initialisation:}\\
$\alpha_0,\beta_0,\theta_0$ \tcp*{contrast, brightness, bias coeffs}
$\lambda_\alpha,\lambda_\beta,\lambda_\theta$ \tcp*{regularisation}
$\eta,\;T$ \tcp*{learning rate and iterations}

\For{$t\gets1$ \KwTo $T$}{
  $I'_0 \gets \alpha I_0 + \beta + B_\theta(I_0,I_0)$\;

  $\mathcal L_{\text{sim}} \gets 1-\text{SSIM}(I'_0,I_r)$\;
  $\mathcal L_{\text{reg}} \gets
        \lambda_\alpha(\alpha-1)^2+
        \lambda_\beta\beta^2+
        \lambda_\theta\lVert\theta\rVert_2^2$\;

  $\mathcal L_{\text{tot}} \gets
        \mathcal L_{\text{sim}}+\mathcal L_{\text{reg}}$\;

  $\varepsilon \gets \varepsilon-\eta
        \nabla_{\!\varepsilon}\mathcal L_{\text{tot}},\;
        \varepsilon=[\alpha,\beta,\theta]$\;

  \If{$\alpha\notin[\alpha_l,\alpha_h]\ \lor\
      \beta\notin[\beta_l,\beta_h]$}{
      $\alpha\gets\text{clip}(\alpha,\alpha_l,\alpha_h)$\;
      $\beta \gets\text{clip}(\beta ,\beta_l ,\beta_h )$\;
  }
  \Else{
      \Continue\;
  }
}
\Return{$(\alpha^\ast,\beta^\ast,\theta^\ast)$}
\caption{Relighting‑mapping optimisation}
\label{algrelighting}
\end{algorithm}

In this study, we proposed an optimization method to adjust the contrast and brightness of the 3D rendered character to maximize its similarity with the background image while preserving its original texture. As shown in Alg.~\ref{algrelighting}, we adjust the original image \( I_{\text{o}} \) for contrast and brightness, obtaining the image \( I_{\text{o}}' \) with altered lighting characteristics, and match it with the given real image \( I_{\text{r}} \). The adjustment process is formulated as follows:

\begin{equation}
    I_{\text{o}}' = \alpha^*a \cdot I_{\text{o}} + \beta^* + \theta^*    
\end{equation}
where, \( \alpha \) is the contrast adjustment factor, \( \beta \) is the brightness adjustment factor, and \( \theta \) is the bias coefficient. The bias coefficient is used to prevent excessive differences between the rendered image and the original texture. Our objective is to optimize the parameters \( \alpha \), \( \beta \), and \( \theta \) to minimize the total loss function, which consists of two main components: similarity loss and regularization loss.

To ensure that the rendered character matches the real background image, we use the Structural Similarity Index (SSIM) to quantify the similarity between the rendered image and the real image. Specifically, we first compute the similarity loss defined as \( 1 - f_s(I_{\text{o}}, I_{\text{r}}) \), ensuring that the rendered image \( I_{\text{o}}' \) is structurally similar to the real image \( I_{\text{r}} \). We then calculate the regularization terms to control the magnitude of the parameters \( \alpha \), \( \beta \), and \( \theta \). These regularization terms prevent overfitting by penalizing large parameter values. The regularization terms apply penalties to overly large parameter values, ensuring model stability. Ultimately, the total loss function is the sum of the similarity loss \(\mathcal{L}_{sim} \) and the regularization loss \(\mathcal{L}_{reg}\).

\begin{equation}
\begin{aligned}
\mathcal{L}_{tot.} &= \mathcal{L}_{sim} + \mathcal{L}_{reg},
\end{aligned}
\end{equation}
where
\begin{equation}
\begin{aligned}
\left\{\begin{matrix}
\mathcal{L}_{sim} =  1 - f_s(I_{\text{o}}, I_{\text{r}}), \\[2ex]
    \mathcal{L}_{reg} = \lambda_\alpha(\alpha-1)^2 +\lambda_\beta\beta^2 + \lambda_\theta\|\theta\|_2^2.
\end{matrix}\right.
\end{aligned}
\end{equation}

We use the Adam optimizer to optimize the parameters \( \alpha \), \( \beta \), and the bias coefficient \( \theta \) to minimize the total loss function. The proposed optimization method adjusts brightness and contrast of rendered image \( I_{\text{o}}' \) via linear transformations while employing a nonlinear correction to preserve texture features, to mache the real image \( I_{\text{r}} \).

\subsection{Joint optimization mechanism}
\label{Joint}

In this part, we introduce the joint loss framework proposed in this paper to address the challenges of high level of intra-class variety and realistic factors.




\subsubsection{\textbf{Detection Loss}}The detection loss \( \mathcal{L}_{\text{det}} \) operates by prioritizing bounding box predictions that achieve both high localization accuracy and confident classification, as defined in Eq.~\ref{det_loss} and Eq.~\ref{max_iou}.

\begin{equation}
\mathcal{L}_{\text{det}}(p) = \text{Conf}_{i^*}(p)
\label{det_loss}
\end{equation}
where
\begin{equation}
i^* = \text{arg}\max(s_{i, j} \mid \text{IoU}(b_{i, j}, b_{\text{gt}, j}) \ge \tau ,\, l_{i, j} = 1)
\label{max_iou}
\end{equation}

Here, \( \mathcal{L}_{\text{det}} \) evaluates predictions based on spatial alignment and semantic relevance. For each candidate box \( b_{i,j} \), the loss ensures geometric consistency by verifying if its IoU with the ground truth \( b_{\text{gt},j} \) exceeds a threshold \( \tau \). Simultaneously, the binary indicator \( l_{i,j} \) confirms its validity as a positive sample. Among all valid candidates, the box with the highest IoU is selected, and its associated confidence score \( Conf_{i^*}(p) \) is used as the detection loss. The loss focuses the adversarial signal on the most spatially aligned and semantically relevant prediction, thereby enhancing the attack effectiveness while maintaining gradient stability. Since the IoU distribution is relatively smooth and reflects spatial alignment, this approach results in a stable gradient flow during optimization. In this study, both 2D detection loss \( \mathcal{L}_{\text{det}}^{\text{patch}} \) and 3D detection loss \( \mathcal{L}_{\text{det}}^{\text{real}} \) are calculated using this method.

\subsubsection{\textbf{Total variation loss (TV loss)}}

The total variation loss encourages spatial smoothness in the adversarial patch by penalizing abrupt pixel intensity changes. It is defined as:  

\begin{equation}
\mathcal{L}_{tv}(p) = \sum_{i,j} \Big( | p_{i+1, j} - p_{i,j} | + | p_{i, j+1} - p_{i,j} | \Big),
\end{equation}  
where \( p_{i,j} \) represents the pixel intensity at position \( (i, j) \). The first term enforces smoothness along the horizontal direction, while the second term does so along the vertical direction. By minimizing TV loss, the optimization discourages high-frequency noise and unnatural artifacts, ensuring that the adversarial patch remains visually consistent and less perceptible.

\subsubsection{\textbf{Patch Optimize}}

During adversarial sample training, the detection model is treated solely as a fixed closed-box for loss calculation and gradient propagation during patch optimization. Inspired by stochastic depth~\cite{huang2016deep,huang2023t}, we employ the \textit{shakedrop regularization} mechanism that randomly adjusts both the forward-output and backward-gradient flows in the residual block~\cite{yamada2018shakedrop}. Specifically, in the forward pass, we combine the identity mapping with the stacked layers’ output to generate new feature representations. For an input feature \(\ x_{\text{in}}\) and the stacked layers’ output \(H(x_{\text{in}})\), the fusion rule is formulated as:
\[
x_{\text{out}} = x_{\text{in}} + (\gamma_1 + \omega - \gamma_1 \cdot \omega)\,H(x_{\text{in}}),
\]
where \(\gamma_1\) is a Bernoulli random variable with \(P(\gamma_1 = 1) = p_s\), and \(\omega\) is sampled from a continuous uniform distribution \(\omega \sim U(1 - k,\,1 + k)\). Here, \(k\) is a predefined constant.

During backpropagation, let \(\mathcal{L}\) denote the loss function. The gradient adjustment follows the chain rule:
\[
\mathcal{G}_{x_{\text{in}}} 
= \mathcal{G}_{x_{\text{out}}} \,\cdot\,
\Bigl(1 \;+\; (\gamma_2 + \omega - \gamma_2 \cdot \omega)
\,\frac{\partial H}{\partial x_{\text{in}}}\Bigr),
\]
where \(\gamma_2\) is another Bernoulli variable drawn from the same distribution as \(\gamma_1\). Through stochastic coefficient fusion in both forward propagation and gradient computation, this approach injects perturbations that enhance the generalization capability of adversarial samples.

The random perturbations effectively create multiple virtual model variants without requiring additional model training. Although the perturbation process does not directly update model parameters, it diversifies the computational pathways, enabling attackers to produce highly transferable adversarial samples using only a single glass-box model.

Building on these benefits, the adversarial patch is iteratively refined. At each step \( i \), the next-step rendering of the human texture is defined as:

\begin{equation}
R^{\text{hum}}(p_{i+1}) = g \Big(p_{i} + \epsilon \cdot \text{sign}\big(\nabla_p \mathcal{L}(f_h(x^{\text{adv}}(p)), y)\big)\Big),
\label{eq3}
\end{equation}
where \( f_h : [-1, 1]^d \to \mathbb{R}^C \) is a pre-trained human detection model mapping the adversarial image \( x^{\text{adv}} \) to class probabilities. The function \( g(\cdot) \) renders the adversarial patch onto the 3D human model, ensuring its adaptation to the surfaces geometry.

The total adversarial loss for minimization is

\begin{equation}
\mathcal{L}(f(x^{\text{adv}}(p)), y)=\mathcal{L}_{\text{total}} = \mu_1 \cdot \mathcal{L}_{\text{det}}^{\text{patch}} + \mu_2 \cdot \mathcal{L}_{\text{det}}^{\text{real}} + \mu_3 \cdot \mathcal{L}_{\text{tv}}
\label{total}
\end{equation}

where the coefficients $\mu_1$, $\mu_2$, and $\mu_3$ are loss weights that balance the importance of 2D attacking effectiveness, 3D attacking effectiveness, and visual naturalness of the patch. Their values are set following the weighting strategy used in~\cite{huang2023t}, and we maintain fixed ratios across experiments. The weighted loss formulation allows us to flexibly emphasize different optimization goals depending on specific threat models or deployment scenarios.

The adversarial patch \( p_{i} \) is optimized to reduce the detection model's performance by applying a gradient-based update. \( \epsilon \cdot \text{sign}(\nabla_p \mathcal{L}(f(x^{\text{adv}}(p)), y)) \) adjusts the patch in the direction that maximally alters the model’s prediction, where \( \nabla_p \mathcal{L} \) is the gradient of the loss function with respect to the patch, \( \epsilon > 0 \) is the step size, and \( \text{sign}(\cdot) \) extracts the gradient's sign.

\section{Experiments}
\begin{figure*}[htbp!]
\centering
\includegraphics[width=15 cm]{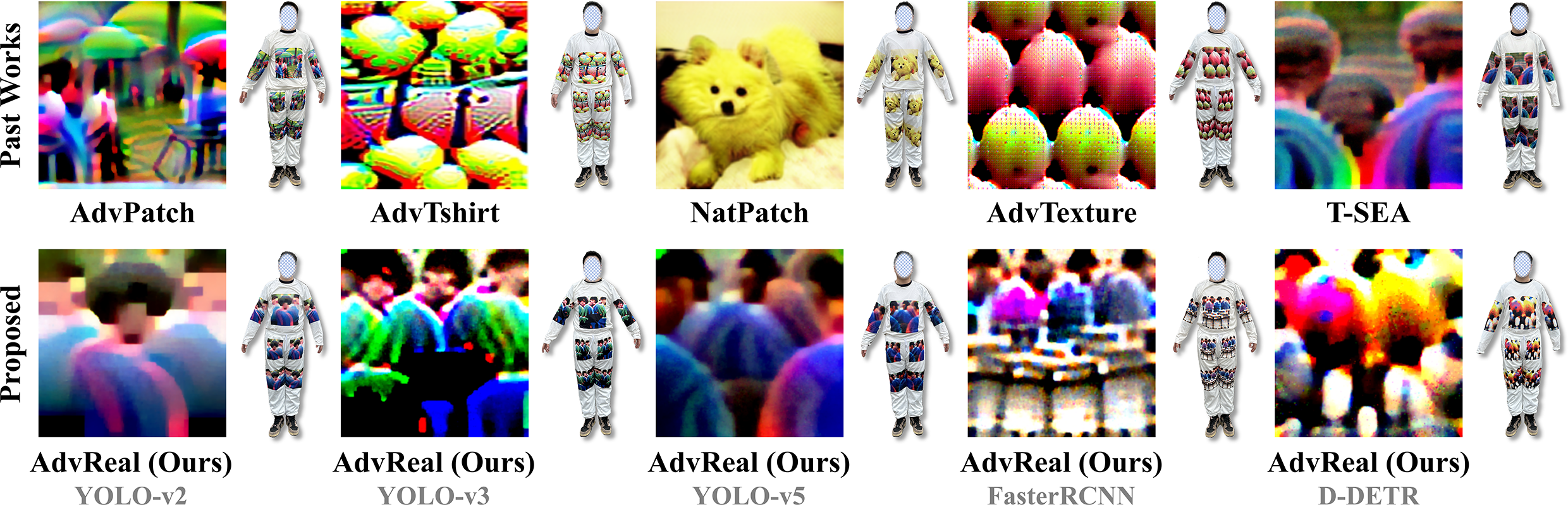}
\caption{Adversarial patches of other advanced algorithms and ours.}
\label{patch}
\vspace{-4 mm} 
\end{figure*}

In this section, we conduct extensive experiments to verify the effectiveness of the proposed AdvReal method. We first describe the experimental setup and then compare it with SOTA methods in both the digital and physical world.

\subsection{Experiments Setup}
\subsubsection{\textbf{Dataset}}

In the \textbf{Patch Adversarial Module (2D adversarial)}, we employ the INRIA Person dataset~\cite{dalal2005histograms} to train and test our adversarial patch, which comprises 462 training images and 100 test images.

In the \textbf{Realistic Adversarial Module (3D adversarial)}, we randomly selected 562 traffic scene images from the nuScenes dataset~\cite{caesar2020nuscenes}, captured from six camera perspectives (Front, Front Right, Front Left, Back Right, Back Left, and Back), to serve as backgrounds for 3D rendering in evaluating our adversarial patches. Among these, 450 images were captured under adequate lighting and 112 under poor lighting. The daytime and nighttime images were randomly divided into a training set of 462 images and a test set of 100 images.

\subsubsection{\textbf{Victim Detector}}

We perform adversarial training and attacks on current mainstream object detectors, including single-stage, two-stage, and transformer-based architectures (shown in Tab.~\ref{detector}). The models are pre-trained on MS COCO dataset~\cite{lin2014microsoft}. In particular, we conduct closed-box attack tests on SOTA detectors (YOLO-v8, v11, v12) in the physical world for the first time. Floating Point Operations (FLOPs) quantify the computational workload required by a model during inference and serve as an indicator of the model’s complexity. In general, models with higher complexity tend to exhibit improved robustness and generalization capabilities.

\begin{table}[h]
\centering
\caption{Victim Detector for evaluating the effectiveness of adversarial patches, including single-stage, two-stage, and transformer-based models.}
\label{detector}
\resizebox{12cm}{!}{
\renewcommand{\arraystretch}{1.15} 
\begin{tabular}{ccccc}
\hline
Detector    & Features         &     FLOPs        & Years & Cites \\ \hline
YOLO-v2     & \multirow{3}{*}{Single-stage} &  5.58 &   2017  & \cite{redmon2017yolo9000}     \\
YOLO-v3     &                               &  193.89 &   2018  & \cite{redmon2018yolov3incrementalimprovement}     \\
YOLO-v5     &                               &  7.70  &   2020  & \cite{yolov5}     \\
\cline{2-5}
YOLO-v8     & \multirow{3}{*}{\begin{tabular}[c]{@{}c@{}}Single-stage\\(SOTA)\end{tabular}} &  8.7  & 2023  & \cite{yolov8_ultralytics}     \\
YOLO-v11    &                               &  6.50  &   2024  & \cite{yolo11_ultralytics}     \\
YOLO-v12    &                               &  6.50 &   2025  & \cite{tian2025yolov12}     \\ \cline{2-5} 
Faster-RCNN & Two-stage                  & 180.00 &   2016  & \cite{ren2016fasterrcnnrealtimeobject}     \\ \cline{2-5} 
D-DETR      & Transformer-based             &  78.00 &   2021  & \cite{ddetr}     \\ \hline
\end{tabular}
}
\vspace{-4 mm} 
\end{table}

\subsubsection{\textbf{Evaluation Metric}}



In the evaluation framework, we define an attack as successful if the Intersection over Union (IoU) between the detection box and the ground-truth box is less than 0.5, or if the detection confidence score is below $0.5$. We use multiple quantitative metrics to comprehensively evaluate the effectiveness of algorithm-generated adversarial patches, as follows:

\begin{itemize}
    \item \textbf{Attacking Success Rate (ASR):} The percentage of adversarial samples that cause the detector to fail, including false detection and missed detection. It is defined as:
    \begin{equation}
        \mathrm{ASR} = \frac{N_{\text{fail}}}{N_{\text{total}}} \times 100\%
    \end{equation}
    where $N_{\text{fail}}$ is the number of images where the attack leads to a missed detection, and $N_{\text{total}}$ is the total number of adversarial samples.

    \item \textbf{Precision:} The ratio of correctly predicted positive detections to all predicted positives. Lower precision indicates that the detector is more effectively misled by the adversarial patch. It is computed as:
    \begin{equation}
        \mathrm{Precision} = \frac{TP}{TP + FP}
    \end{equation}
    where $TP$ denotes true positives, which refer to correct detections that match ground-truth objects. $FP$ refers to false positives, meaning incorrect detections without corresponding ground-truth matches.

    \item \textbf{Recall:} The ratio of correctly detected ground-truth targets to the total number of ground-truth targets. Lower recall signifies that more objects are missed due to the attack. It is computed as:
    \begin{equation}
        \mathrm{Recall} = \frac{TP}{TP + FN}
    \end{equation}
    where $FN$ indicates false negatives, which are ground-truth objects that are missed by the detector.

    \item \textbf{F1-Score:} The harmonic mean of precision and recall, providing a balanced measure of detection quality. It is calculated as:
    \begin{equation}
        \mathrm{F1} = 2 \cdot \frac{\mathrm{Precision} \cdot \mathrm{Recall}}{\mathrm{Precision} + \mathrm{Recall}}
    \end{equation}

    \item \textbf{Average Confidence (AC):} The average confidence score of the detector’s predictions. A lower AC suggests that the detector is less certain about its outputs under attack. It is defined as:
    \begin{equation}
        \mathrm{AC} = \frac{1}{N} \sum_{i=1}^{N} s_i
    \end{equation}
    where $s_i$ is the confidence score of the $i$-th detection, and $N$ is the number of valid predictions.
\end{itemize}

\subsubsection{\textbf{Experimental Details}}

To improve reproducibility and readability, we summarize the experimental setup and training hyperparameters in Tab.~\ref{setup}. During the digital train and evaluation, the distance from 3D human model to camera is limited in $1$ to $4$ meters. All rendered human models are at random angles and positions, and follow the principle of time-space mapping in Sec.~\ref{time_space}.

In non-rigid surfaces modeling module, $\sigma, \gamma, \rho$ in eq.~\ref{sigma} is set as $0.8$, and $\gamma, \rho$ in eq.~\ref{C} are set as $0.01,0.2$ respectively. The the total number of training rounds is set to $800$. The initial training patch size is set as $300 \times 300$, the input image is $416 \times 416$. We choose Adam as the optimizer and the learning rate is set as $0.01$. All experiments are conducted with 16G NVIDIA GeForce RTX 4080 GPU.
\begin{table*}[h]
\centering
\caption{Experimental environment and parameters settings.}
\label{setup}
\resizebox{14cm}{!}{
\begin{tabular}{llll}
\toprule
\textbf{Setting} & \textbf{Value} & \textbf{Setting} & \textbf{Value} \\
\midrule
GPU & NVIDIA RTX 4080 (16GB) & Image Resolution & $416 \times 416$ \\
Patch Size (Init) & $300 \times 300$ & Training Epochs & 800 \\
Batch Size & 8 & Optimizer & Adam \\
Learning Rate & 0.01 & Patch Initialization & Random Uniform \\
Loss Weights & $\mu_1=1.0$, $\mu_2=1.0$, $\mu_3=2.5$ & Random Seed & 42 \\
Camera Distance & 1 to 4 meters & Viewpoint Sampling & Random angle and position \\
Non-Rigid Params & $\sigma=0.8$, $\gamma=0.01$, $\rho=0.2$ & Rendering Principle & Time-space mapping (Sec.~\ref{time_space}) \\
\bottomrule
\end{tabular}
}
\end{table*}

\subsubsection{\textbf{Baseline}}



We compared our method \textit{AdvReal} with both classic and state-of-the-art (SOTA) adversarial methods. The classic methods include \textit{AdvPatch}~\cite{brown2017adversarial}, \textit{AdvTshirt}~\cite{xu2020adversarial}, and \textit{NatPatch}~\cite{hu2021naturalistic}, while the SOTA methods comprise \textit{AdvTexture}~\cite{hu2022adversarial}, \textit{AdvCaT}~\cite{hu2023physically}, and \textit{T-SEA}~\cite{huang2023t}. In addition, we assess \textit{White}, \textit{Gray}, and \textit{Random Noise} patches as control groups. For fairness, all baseline methods are evaluated using adversarial patches trained on YOLOv2 or Faster R-CNN, which are the original glass-box detectors used in their respective papers. Few of baseline methods support adversarial patch generation for more recent or structurally different detectors such as YOLOv5 or D-DETR. In contrast, our method AdvReal supports adversarial training on multiple detectors (e.g., YOLOv2, YOLOv3, YOLOv5, Faster R-CNN, D-DETR), enabling broader cross-model evaluation and generalization analysis. For fairness, the areas of all adversarial patches on both rendered and real garments are kept consistent across digital and physical experiments. With the exception of AdvCat, which is entirely camouflaged, the remaining patches and their corresponding garments are shown in Fig.~\ref{patch}. 

To comprehensively assess attack performance across continuous video frames and multiple angles, all adversarial patches are printed on long-sleeve tops and trousers. Specifically, six patches are applied to the top—two on the front and back centers and four on the sleeves (one on the front and one on the back of each sleeve)—while eight patches are applied to the trousers, distributed evenly across both legs.

\subsection{Digital Adversarial Attack Experiment}
\subsubsection{\textbf{Comparisons to SOTA Methods}}

\begin{table*}[ht]
\setlength{\tabcolsep}{4pt} 
\renewcommand{\arraystretch}{1.15} 
\centering
\caption{Comparisons with existing detection attack methods. Compared with existing methods, the proposed AdvReal achieves the best performance on both glass-box attack and closed-box attack.}
\resizebox{16cm}{!}{

\begin{tabular}{c|c|ccccccc|c}
\hline
\multirow{2}{*}{\textbf{Method}} & \multicolumn{1}{c|}{\makecell{\textbf{ASR $\uparrow$} \\ \textbf{(Glass-box)}}} & \multicolumn{7}{c|}{\textbf{ASR $\uparrow$ (Closed-box)}} & \multirow{2}{*}{\makecell{\textbf{ASR $\uparrow$} \\ \textbf{(Closed-box Avg)}}} \\
\cline{2-9}
 & \textbf{YOLOv2} & \textbf{YOLOv3} & \textbf{YOLOv5} & \textbf{YOLOv8} & \textbf{YOLOv11} & \textbf{YOLOv12} & \textbf{F-RCNN} & \textbf{D-DETR} &  \\
\hline
White             & 1.08\%  & 0.43\%  & 2.60\%  & 1.95\%  & 5.63\%  & 2.60\%  & 0.22\%  & 3.68\%  & 2.44\%  \\
Gray              & 0.39\% & 0.43\%  & 4.11\%  & 4.98\%  & 6.49\%  & 3.90\%  & 0.22\%  & 6.71\%  & 3.83\%  \\
Noise             & 0.22\%  & 0.87\%  & 5.19\%  & 10.39\% & 4.55\%  & 4.98\%  & 0.43\%  & 3.68\%  & 4.30\%  \\ \hline
AdvPatch          & 33.55\% & 3.46\%  & 51.52\% & \underline{42.64\%} & 22.51\% & 16.45\% & 6.06\%  & 30.30\% & 24.71\%  \\
AdvTshirt         & 24.68\% & 12.77\% & 43.72\% & 35.28\% & \underline{27.92\%} & 15.80\% & 1.52\%  & 31.39\% & 24.06\%  \\
NatPatch          & 6.06\%  & 4.98\%  & 16.23\% & 10.39\% & 8.87\%  & 4.11\%  & 0.22\%  & 13.20\% & 8.29\%  \\
AdvTexture        & 55.63\% & \underline{75.11\%} & \underline{69.05\%} & 28.35\% & 22.29\% & 19.70\% & 13.64\% & 52.60\% & \underline{40.11\%}  \\
AdvCaT            & 0.00\%  & 0.00\%  & 0.00\% & 0.65\%  & 0.43\%  & 0.87\%  & 0.00\%  & 0.22\%  & 0.31\%  \\
T-SEA             & \underline{61.90\%} & 38.10\% & 50.65\% & 39.39\% & 23.59\% & \underline{21.65\%} & \underline{32.25\%} & \underline{51.52\%} & 36.74\%  \\
AdvReal (Ours)    & \textbf{89.39\%} & \textbf{77.92\%} & \textbf{74.46\%} & \textbf{67.32\%} & \textbf{67.53\%} & \textbf{70.13\%} & \textbf{51.08\%} & \textbf{59.09\%} & \textbf{66.79\%}  \\
\hline
\end{tabular}
}
\label{compare}
\end{table*}

In Tab.~\ref{compare}, we compare the proposed \textit{AdvReal} with three non-adversarial textures and seven state-of-the-art adversarial patches. YOLOv2 is employed as the glass-box detector, while seven additional closed-box detectors are used to evaluate the adversarial patches.


In the glass-box comparison, influenced by realistic digital world samples, the ASR of previous adversarial patches dropped considerably compared to the results reported in their original papers, revealing limited generalization under realistic deployment conditions. Moreover, the ASR of the natural-looking methods—NatPatch ($6.06\%$) and AdvCaT ($0.00\%$)—is nearly the three non-adversarial textures. In particular, the $0\%$ ASR observed for AdvCaT across multiple detectors highlights the vulnerability of camouflage-based attacks under real-world constraints. This method relies on high-frequency texture signals that are effective under controlled, close-range, high-resolution settings, but degrade significantly in realistic conditions involving longer viewing distances, varying angles, and lower resolution. Additionally, the natural appearance of the camouflage may inadvertently align with semantic cues of the “person” class, potentially reinforcing the detector's confidence rather than suppressing it. Overall, we attribute the performance drop to training approaches that do not adequately account for real world factors such as body geometry, self-occlusion, and dynamic illumination.

Under the closed-box comparison, AdvReal achieves consistently higher attack success rates than other adversarial methods, including state-of-the-art approaches such as AdvTexture and T-SEA. Notably, when evaluated against the advanced detector YOLOv12, AdvReal achieves an ASR of $70.13\%$, dramatically outperforming T-SEA ($21.65\%$).

It is worth noting that the observed gap in adversarial performance across detectors stems from the architectural discrepancy between the black-box evaluation models and the glass-box detector used for adversarial training. Specifically, as shown in Tab.2, all adversarial patches are generated through adversarial training on YOLOv2. Since YOLOv3 shares a similar one-stage architecture with YOLOv2, the patches transfer effectively, yielding a high ASR of 77.92\%. Compared to YOLO series detectors, the two-stage detectors and the transformer-based detectors adopt fundamentally different detection pipelines~\cite{zhang2021mining}. For example, Faster R-CNN relies on two-stage region proposal networks, while D-DETR uses attention-based token representations. The result shows that, adversarial patches against with specific architecture have significantly lower attack success rates on heterogeneous detectors~\cite{nguyen2025survey}.


To further validate AdvReal’s adversarial performance, we evaluated the ASR of various adversarial patches under different IoU thresholds. In this experiment, both YOLOv2 and Faster R-CNN are used as glass-box detectors, meaning that the adversarial patches are generated and tested under the same detector architecture to measure their optimal attack capacity. As shown in Tab.~\ref{IOU}, AdvReal achieves higher ASRs than other attack methods when using both YOLOv2 and Faster R-CNN as detectors. Under the YOLOv2 detector, AdvReal attains an ASR of $45 \%$ at an IoU threshold of $0.1$—surpassing T-SEA’s $29.00 \%$ ASR at an IoU threshold of $0.5$. With the Faster R-CNN detector, AdvReal’s ASR approaches $100 \%$ at an IoU threshold of $0.5$, significantly exceeding T-SEA’s $67.10 \%$. The results indicate that AdvReal can achieve nearly complete suppression of detection outputs when evaluated in a white-box setting, regardless of the detector type.

\begin{table*}[ht!]
\centering

\caption{ASRs (\%) under different IoU thresholds, using YOLOv2 and Faster R-CNN as glass-box detectors and victim detectors}

\resizebox{16cm}{!}{
\setlength{\tabcolsep}{4pt} 
\renewcommand{\arraystretch}{1.15} 
\begin{tabular}{c|ccccc|ccccc}
\hline
\multirow{2}{*}{\textbf{Method}} 
& \multicolumn{5}{c|}{\textbf{YOLOv2 (Glass-box)} $\uparrow$} 
& \multicolumn{5}{c}{\textbf{Faster-RCNN (Glass-box)} $\uparrow$} \\ 
\cline{2-11}
& \textbf{IoU=0.1} & \textbf{IoU=0.3} & \textbf{IoU=0.5} & \textbf{IoU=0.7} & \textbf{IoU=0.9} 
& \textbf{IoU=0.1} & \textbf{IoU=0.3} & \textbf{IoU=0.5} & \textbf{IoU=0.7} & \textbf{IoU=0.9} \\
\hline
White
& 0.43\% & 0.43\% & 0.65\% & 6.49\%  & 70.78\% 
& 0.22\% & 0.22\% & 0.22\% & 2.16\%  & 53.90\% \\
Gray         
& 0.00\% & 0.00\% & 0.00\% & 7.79\%  & 80.74\% 
& 0.00\% & 0.00\% & 0.22\% & 1.30\%  & 68.18\% \\
Random Noise      
& 0.22\% & 0.00\% & 0.22\% & 8.87\%  & 78.79\%
& 0.22\% & 0.43\% & 0.43\% & 2.38\%  & 62.77\% \\ \hline
AdvPatch     
& 12.77\% & 12.99\% & 14.07\% & 39.83\% & 88.96\% 
& 0.87\%  & 1.08\%  & 6.06\% & 22.94\% & 91.77\% \\
AdvTshirt    
& 11.26\% & 11.26\% & 11.26\% & 29.65\% & 91.13\% 
& 0.43\%  & 0.43\%  & 1.52\% & 8.87\%  & 87.45\% \\
NatPatch     
& 1.52\%  & 1.52\%  & 1.52\%  & 10.61\% & 83.33\% 
& 0.22\%  & 0.22\%  & 0.22\% & 1.95\%  & 80.74\% \\
AdvTexture   
& 25.11\% & 25.54\% & 25.76\% & 60.61\% & 96.97\% 
& \underline{2.38\%} & \underline{3.68\%} & 13.64\% & 41.56\% & \underline{98.05\%} \\
AdvCaT       
& 0.00\%  & 0.00\%  & 0.00\%  & 3.25\%  & 74.46\% 
& 0.00\%  & 0.00\%  & 0.00\%  & 0.22\%  & 52.38\% \\
T-SEA         
& \underline{29.00\%} & \underline{30.74\%} & \underline{38.10\%} & \underline{78.14\%} & \underline{98.27\%} 
& 0.22\% & 0.43\% & \underline{67.10\%} & \underline{89.18\%} & \textbf{100.00\%} \\ 
AdvReal (Ours) 
& \textbf{45.02\%} & \textbf{59.31\%} & \textbf{86.80\%} & \textbf{98.27\%} & \textbf{100.00\%} 
& \textbf{6.71\%} & \textbf{43.94\%} & \textbf{99.57\%} & \textbf{99.78\%} & \textbf{100.00\%} \\
\hline
\end{tabular}
}
\label{IOU}
\vspace{-4 mm} 
\end{table*}

\subsection{Ablation Study}

Tab.~\ref{ablation} presents ablation experiments that systematically examine the contributions of three modules in the AdvReal framework—namely, non-rigid surfaces, realistic matching, and ShakeDrop. The results reveal that each module significantly influences both the attack success rate (ASR) and the perceptual quality, measured by average confidence (AC), of adversarial patches. Specifically, the exclusive use of the non-rigid surfaces module boosts ASR from $87.45\%$ to $91.56\%$ while simultaneously lowering AC from $59.68\%$ to $55.64\%$, thereby enhancing both attack efficacy and visual quality. Similarly, when applied individually, both the realistic matching and ShakeDrop modules improve ASR and reduce AC. Notably, the realistic matching module yields a particularly substantial improvement, raising ASR to $90.48\%$ and reducing AC to $55.39\%$. 

Interestingly, the combined application of non-rigid surfaces and ShakeDrop results in a slight ASR decrease to $87.01\%$, yet it substantially improves perceptual quality, as evidenced by an AC reduction to $54.28\%$, indicating a synergistic benefit. Ultimately, the full integration of all three modules delivers the best performance, achieving an ASR of $93.72\%$ and an AC of $53.17\%$. The results confirm the cumulative impact of each component and highlight the framework’s ability to balance attack strength with visual imperceptibility.

\begin{table}[t]
\centering 
\caption{Ablation experiment results}
\resizebox{8.6cm}{!}{
\setlength{\tabcolsep}{5pt} 
\renewcommand{\arraystretch}{1.15} 
\begin{tabular}{ccccc}
\hline 
\begin{tabular}[c]{@{}c@{}}Non-rigid\\ surfaces\end{tabular} & \begin{tabular}[c]{@{}c@{}}Realistic\\ Matching\end{tabular} & \begin{tabular}[c]{@{}c@{}}Shakedrop \\ Regularization\end{tabular} & ASR $\uparrow$ & AC $\uparrow$ \\
\hline 
 &  &  & 87.45\% & 59.68\% \\ 
 \hline 
\checkmark &  &  & 91.56\% & 55.64\% \\ 
 & \checkmark &  & 90.48\% & 55.39\% \\ 
 &  & \checkmark & 89.18\% & 58.79\% \\ 
 \hline 
\checkmark &  & \checkmark & 87.01\% & 54.28\% \\ 
 & \checkmark & \checkmark & 92.42\% & 55.33\% \\ 
\checkmark &  & \checkmark & 88.31\% & 54.84\% \\ 
\hline 
\checkmark & \checkmark & \checkmark & 93.72\% & 53.17\% \\ 
\hline 
\end{tabular}
}
\label{ablation}
\vspace{-4 mm} 
\end{table}

\subsection{Robustness}

\begin{table}[h]
\centering
\caption{Robustness performance of different methods on YOLOv2}
\resizebox{8.6cm}{!}{
\setlength{\tabcolsep}{4pt} 
\renewcommand{\arraystretch}{1.15} 
\begin{tabular}{c|cccc}
\hline
\multirow{2}{*}{\textbf{Method}} & \multicolumn{4}{c}{\textbf{YOLOv2 (Glass-box)}} \\
\cline{2-5}
 & \textbf{ASR $\uparrow$} & \textbf{Precision $\downarrow$} & \textbf{Recall $\downarrow$} & \textbf{F1-Score $\downarrow$} \\
\hline
AdvPatch       & 26.84\% & 73.16\% & 60.68\% & 60.68\% \\
AdvTshirt      & 20.56\% & 56.03\% & 79.44\% & 65.71\% \\
NatPatch       & 3.46\%  & 62.12\% & 96.54\% & 75.59\% \\
AdvTexture     & 21.86\% & 57.76\% & 78.14\% & 66.42\% \\
T-SEA           & \underline{36.15\%} & \underline{49.50\%} & \underline{63.85\%} & \underline{55.77\%} \\
AdvReal        & \textbf{84.20\%} & \textbf{14.87\%} & \textbf{15.80\%} & \textbf{15.32\%} \\\hline
\end{tabular}
}
\label{robustness}
\vspace{-4 mm} 
\end{table}

When the adversarial patch is partially occluded, its attack efficacy is markedly reduced. To evaluate the robustness of different algorithms under occlusion, we superimpose a $100 \times 100$ pixel gray square at the center of a $300 \times 300$ adversarial patch. Tab.~\ref{robustness} compares the robustness performance of various methods under the YOLOv2 glass-box setting. As shown in the table, our method achieves an ASR of $84.20\%$, significantly outperforming the second-best method, T-SEA, which records $36.15\%$. Moreover, AdvReal attains the highest values across all detection metrics—Precision, Recall, and F1-Score ($14.87\%$, $15.80\%$, and $15.32\%$, respectively)—demonstrating its strong ability to disrupt detection even under partial occlusion. In contrast, other methods such as AdvPatch, AdvTshirt, and AdvTexture record ASRs below $30.00\%$ and yield substantially lower detection metrics. NatPatch, in particular, achieves only a $3.46\%$ ASR, likely due to its prioritization of visual naturalness over adversarial strength.

\begin{table*}[ht!]
\centering
\caption{Transferability of different attack methods in YOLOv8, YOLOv11, and YOLOv12, the glass-box is YOLOv2}
\resizebox{16cm}{!}{
\setlength{\tabcolsep}{4pt} 
\renewcommand{\arraystretch}{1.15} 
\begin{tabular}{c|ccc|ccc|ccc}
\toprule
\multirow{2}{*}{\textbf{Method}} & \multicolumn{3}{c}{\textbf{YOLOv8}} & \multicolumn{3}{c}{\textbf{YOLOv11}} & \multicolumn{3}{c}{\textbf{YOLOv12}} \\
\cmidrule(lr){2-4} \cmidrule(lr){5-7} \cmidrule(lr){8-10}
& \textbf{Precision $\downarrow$} & \textbf{Recall $\downarrow$} & \textbf{F1-score $\downarrow$} & \textbf{Precision $\downarrow$} & \textbf{Recall $\downarrow$} & \textbf{F1-score $\downarrow$} & \textbf{Precision $\downarrow$} & \textbf{Recall $\downarrow$} & \textbf{F1-score $\downarrow$} \\
\midrule
AdvPatch & 93.64\% & 57.36\% & 71.14\% & 96.50\% & 77.49\% & 85.95\% & 98.72\% & 83.55\% & 90.50\% \\
AdvTshirt & 96.76\% & 64.72\% & 77.56\% & 97.65\% & 72.08\% & 82.94\% & 98.98\% & 84.20\% & 90.99\% \\
NatPatch & 97.41\% & 89.61\% & 93.35\% & 97.91\% & 91.13\% & 94.39\% & 99.11\% & 95.89\% & 97.47\% \\
AdvTexture & 96.50\% & 71.65\% & 82.24\% & 97.03\% & 77.71\% & 86.30\% & 98.93\% & 80.30\% & 88.65\% \\
AdvCaT & 98.08\% & 99.35\% & 98.71\% & 98.50\% & 99.57\% & 99.03\% & 98.92\% & 99.13\% & 99.03\% \\
T-SEA-v2 & \underline{83.58\%} & \underline{60.61\%} & \underline{70.26\%} & \underline{94.89\%} & \underline{76.41\%} & \underline{84.65\%} & \underline{98.10\%} & \underline{78.35\%} & \underline{87.12\%} \\
AdvReal (Ours) & \textbf{56.55\%} & \textbf{32.68\%} & \textbf{70.26\%} & \textbf{66.37\%} & \textbf{32.47\%} & \textbf{43.60\%} & \textbf{82.63\%} & \textbf{29.87\%} & \textbf{43.88\%} \\
\bottomrule
\end{tabular}
}
\label{tab:performance}
\end{table*}

\begin{figure}[h]
\centering
\includegraphics[width=8.6 cm]{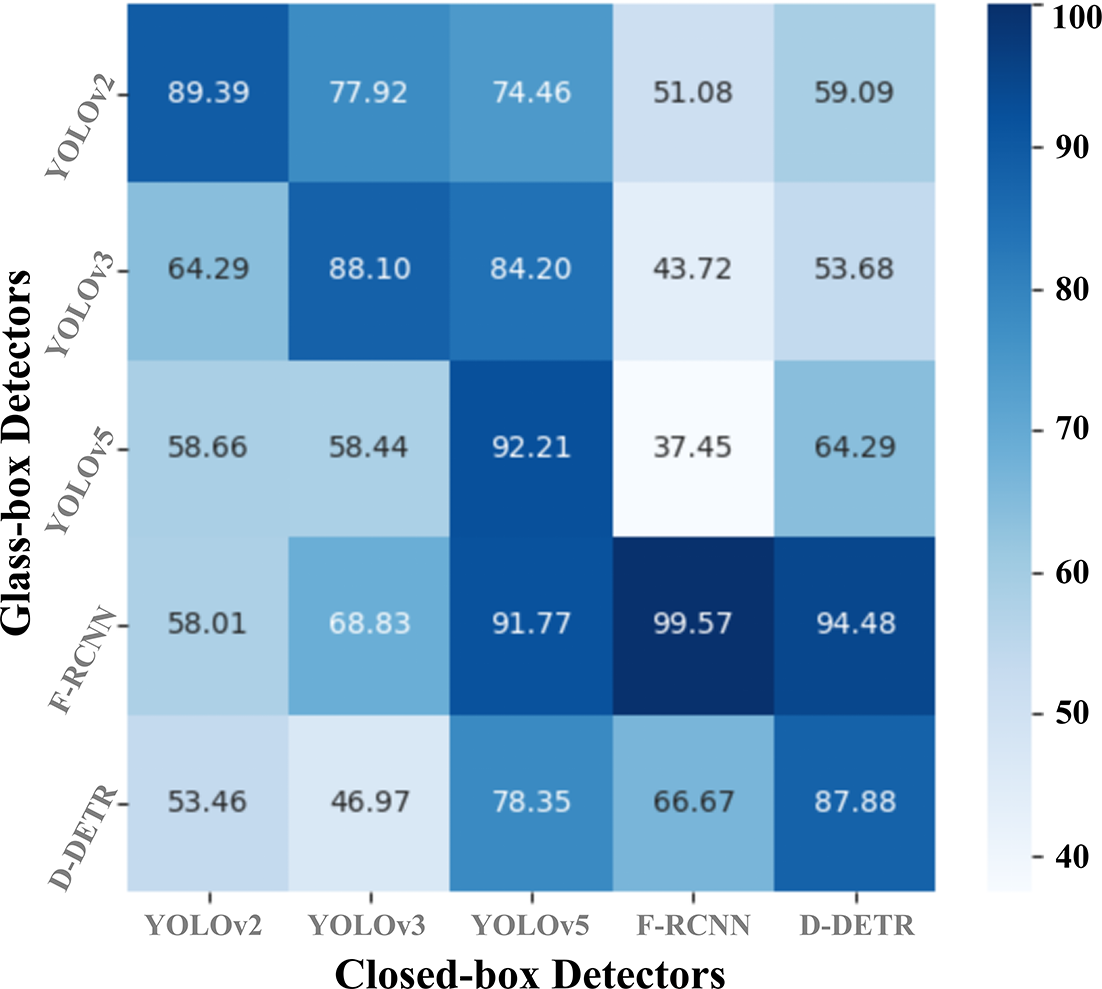}
\caption{ASRs with different method and IOU thresholds in digital world. The confidence threshold is set as $0.5$.}
\label{confusion}
\vspace{-6 mm} 
\end{figure}

\subsection{Transferability}

\subsubsection{\textbf{Transfer confusion matrix}}
As illustrated in Fig.~\ref{confusion}, we optimized adversarial patches on various detectors to verify the transferability of AdvReal and evaluated their ASRs under both glass-box and closed-box settings. In glass-box tests, AdvReal consistently achieved the highest ASRs—exceeding $90 \%$ for YOLOv5 and Faster R-CNN ($92.21 \%$ and $99.57 \%$, respectively). Although the patch trained on the single-stage YOLO algorithm experienced a significant ASR drop when transferred to the two-stage F-RCNN, it still maintained an ASR above $35.00 \%$. This transferability confirms that AdvReal remains effective under closed-box conditions.

\subsubsection{\textbf{Transfer performance across confidence threshold}}
\begin{figure*}[ht!]
\centering
\includegraphics[width=16 cm]{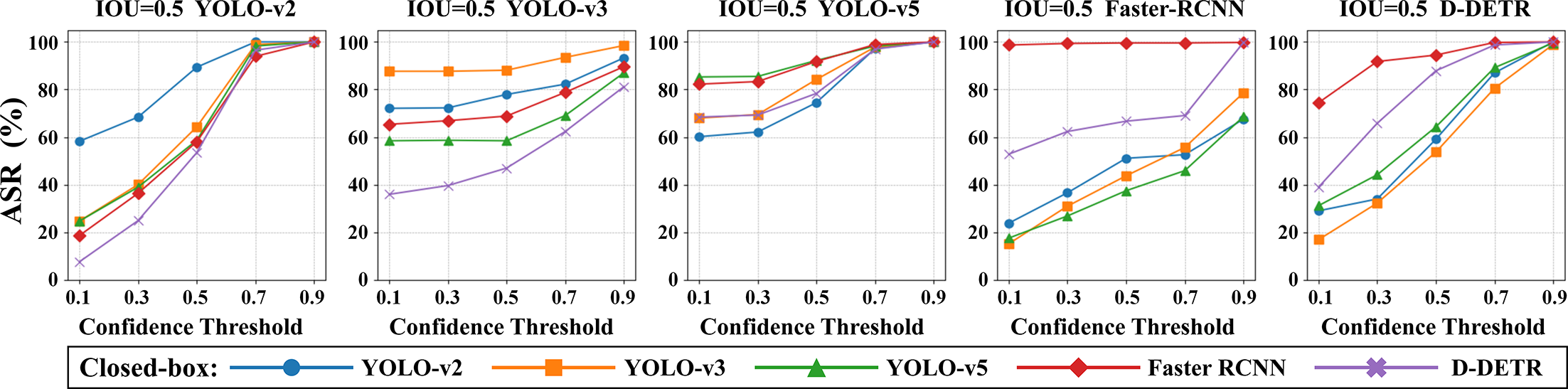}
\caption{ASRs with different detectors and confidence thresholds in digital world. The glass-box detector used for adversarial training is marked on the top of each sub-image, and lines of different colors represent the corresponding detectors used for evaluation.}
\label{confidence}
\vspace{-4 mm} 
\end{figure*}

We evaluated the ASRs on various closed-box detectors at different confidence thresholds. As shown in Fig.~\ref{confidence}, for the YOLOv2 and YOLOv3 detector, adversarial patches trained under the glass-box setting consistently outperformed those trained under the closed-box setting. At a confidence level of $0.7$, both glass-box and closed-box patches achieved nearly $100\%$ ASR for YOLOv2. For the YOLOv5 detector, both glass-box and closed-box ASRs exceeded $60\%$ even at a low confidence threshold of $0.1$. In the Faster R-CNN detector, the glass-box ASR remained around $100\%$ across confidence thresholds from $0.1$ to $0.9$. However, due to structural differences relative to the single-stage YOLO and transformer-based D-DETR architectures, the closed-box ASRs remained relatively low until a confidence level of $0.7$ was reached. In contrast, for the Deformable-DETR detector, the glass-box ASR was lower than the closed-box ASR observed for Faster R-CNN, which can be attributed to the transformer-based architecture of Deformable-DETR that provides greater robustness than other closed-box detectors.

\subsubsection{\textbf{Mobility on advanced detectors}}
To further verify the transferability of AdvReal to the latest detectors, we evaluated various adversarial attack methods on three state-of-the-art (SOTA) detectors: YOLOv8 (2023s), YOLOv11 (2024s), and YOLOv12 (2025s). As shown in Tab.~\ref{tab:performance}, the results indicate that AdvReal consistently yields significantly lower Precision, Recall, and F1-Scores across all detectors, highlighting its superior effectiveness in degrading object detection performance. Specifically, on YOLOv8, AdvReal achieved Precision, Recall, and F1-Score values of 56.55\%, 32.68\%, and 70.26\%, respectively, outperforming all other methods. Similarly, on YOLOv11 and YOLOv12, AdvReal demonstrated outstanding performance, attaining Precision values of 66.37\% and 82.63\%, Recall values of 32.47\% and 29.87\%, and F1-Scores of 43.60\% and 43.88\%, respectively. In contrast, methods such as AdvPatch, AdvTshirt, and AdvTexture performed substantially worse, particularly in the Recall and F1-Score metrics. Notably, although AdvCaT performed reasonably well in Precision and F1-Score, its Recall value was exceptionally high (nearly 99\%), indicating a weaker disruptive capability against the detection models. Overall, AdvReal exhibited superior transferability across different detectors, confirming its robustness and generalization in cross-model adversarial attacks.

\subsection{Visual Quality and Detectability}
\subsubsection{Visual quality}

To quantitatively evaluate the visual quality of different adversarial patches, we employ the PaQ-2-PiQ framework~\cite{ying2020patches}, a no-reference image quality prediction model trained on a large-scale database of real-world distortions and human judgments. This model outputs visual quality scores that are highly correlated with subjective perception, providing a reliable metric for assessing patch naturalness.

\begin{table}[h]
\centering

\caption{Comparison of patch visual quality using PaQ-2-PiQ}
\resizebox{14cm}{!}{
\begin{tabular}{ccccccc}
\hline
\multicolumn{1}{l}{}  & \textbf{AdvPatch} & \textbf{AdvTexture} & \textbf{AdvTshirt} & \textbf{NatPatch} & \textbf{TSEA} & \textbf{AdvReal (Ours)} \\ \hline
\textbf{Visual Score $\uparrow$} & 67.37             & 76.93               & 72.36              & 62.15             & 57.91            & 61.94                   \\
\textbf{Category}     & Good              & Good                & Good               & Fair              & Poor             & Fair                    \\ \hline
\end{tabular}
\label{visual_quality}
}
\end{table}

As summarized in Tab.~\ref{visual_quality}, we evaluate six representative patch types and categorize them based on their perceptual scores. AdvTexture, AdvPatch, and AdvTshirt receive the highest scores and are rated as “Good”, reflecting their visually pleasing appearance or strong integration into background textures. In contrast, T-SEA and NatPatch obtain relatively low scores due to unnatural surface appearance or camouflage overuse.

Our method, AdvReal, achieves a visual quality score of 61.94 and is classified as “Fair”. While this reflects only a moderate level of visual realism, it is a direct result of prioritizing adversarial effectiveness under physical constraints. The optimization of texture alignment, relighting, and geometric adaptation inherently introduces visual artifacts that may reduce perceptual smoothness. Nonetheless, the generated patches remain physically plausible and deployable, striking a practical balance between attack strength and visual subtlety.

\subsubsection{Detectability}
\label{sec_detectability}

To assess the detectability of adversarial patches from a visual security perspective, we employ NAPGuard~\cite{wu2024napguard}, a dedicated framework for identifying naturalistic adversarial patches. NAPGuard distinguishes between aggressive and natural feature components within the patch and applies feature alignment and suppression to robustly detect adversarial content under visual camouflage. We evaluate all patch types using its pretrained model and report detection performance based on the AP@0.5 metric, which measures the ability to correctly localize patch regions.

\begin{table}[h]
\centering

\caption{Patch detectability results using NAPGuard (AP@0.5)} 
\resizebox{14cm}{!}{
\begin{tabular}{ccccccc}
\hline
\multicolumn{1}{l}{}  & \textbf{AdvPatch} & \textbf{AdvTexture} & \textbf{AdvTshirt} & \textbf{NatPatch} & \textbf{TSEA} & \textbf{AdvReal (Ours)} \\ \hline
\textbf{Average Confidence $\downarrow$} & 0.6407   & 0.5952    & 0.4654   & 0.1147   & 0.8225    & 0.6212   \\
\textbf{Detection Rate $\downarrow$}     & 65.02\%  & 63.07\%   & 60.51\%  & 61.37\%  & 71.20\%   & 60.67\%                 \\ \hline
\end{tabular}
\label{detectability}
}
\end{table}

Tab.~\ref{detectability} summarizes the detection results across six representative methods. Patches with high adversarial saliency and minimal visual blending, such as T-SEA, exhibit the highest detectability ($0.8225$). In contrast, NatPatch, which emphasizes visual camouflage, has the lowest detection rate ($0.1147$). AdvReal achieves a moderate detectability score of $0.6212$, which is comparable to AdvPatch ($0.6407$) and AdvTexture ($0.5952$). This reflects the fact that while AdvReal is not specifically designed to evade detection, it also avoids conspicuous or structurally repetitive textures that would make it easily identifiable.

\begin{figure*}[ht!]
\centering
\includegraphics[width=12 cm]{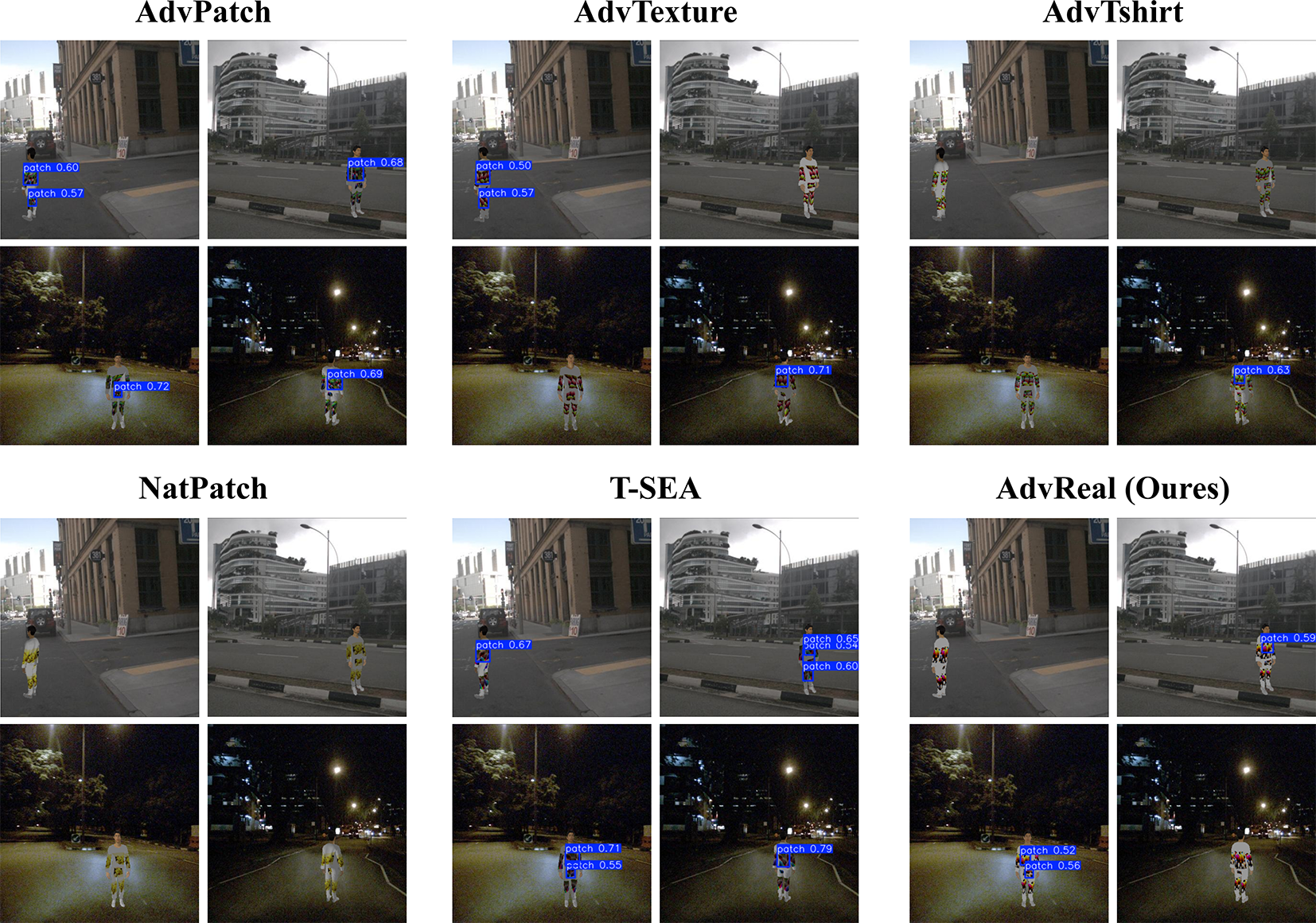}
\caption{NAPGuard detection visualizations. Patches with aggressive features yield stronger activation. AdvReal maintains moderate detectability.}
\label{detectability_fig}
\vspace{-4 mm} 
\end{figure*}

Fig.~\ref{detectability_fig} provides the qualitative examples of patch detection results by NAPGuard. It can be seen that NAPGuard effectively localizes regions in patches with strong adversarial structure, while visually incoherent or low-signal patches evade detection. NatPatch typically triggers minimal activation due to its camouflage-like appearance, whereas T-SEA produces dense and easily localized responses, reflecting its high adversarial saliency but low visual subtlety. AdvReal exhibits consistent but non-dominant activation, suggesting that its adversarial patterns are strong enough to be partially detected, yet not overly distinguishable.

\subsection{Physical Adversarial Attack Experiment}

The physical adversarial attack experiments were conducted in real-world settings, where human subjects wore adversarial shirts and pants with printed patches.

\begin{figure}[h!]
\centering
\includegraphics[width=8.6 cm]{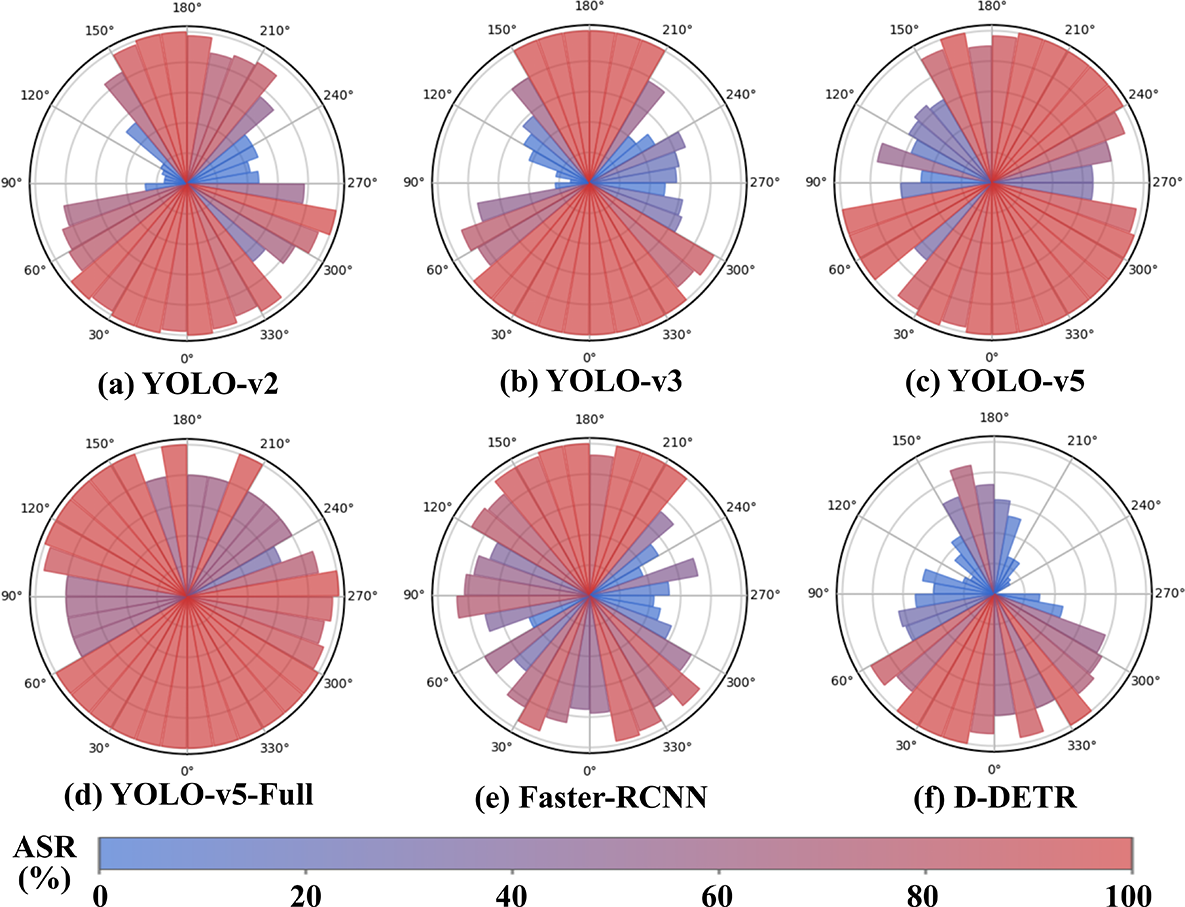}
\caption{ASRs of adversarial patches trained with different glass-box detectors at various angles. All experimental results are obtained by adversarial training and evaluation on the models annotated in the subgraphs.}
\label{angle}
\vspace{-4 mm} 
\end{figure}

\subsubsection{\textbf{All-angle attack}}
To evaluate the robustness of adversarial patches in real-world applications, we recorded videos of subjects wearing AdvReal-patched clothing while performing a full 360° rotation. For each patch, ten videos with different backgrounds were recorded. We assessed performance consistency across different viewpoints by computing the average ASR at $10°$ intervals. As shown in Fig.~\ref{angle}, adversarial patches trained on different detectors exhibit distinct angle-dependent attack effectiveness. The experimental results demonstrate that patches achieve better multi-angle attack performance against the single-stage YOLO glass-box detector. In general, all patches exhibit higher ASRs when facing the camera from the front or back. As the body rotates sideways, the ASR tends to decrease. This drop in performance is attributed to reduced visible patch area and the increased prominence of non-patched human features such as the face, hands, and feet.

\begin{table}[t]
\centering
\caption{Physical‑world ASRs of various methods at different distances under adequate lighting conditions}
\resizebox{7cm}{!}{%
\setlength{\tabcolsep}{4pt}
\renewcommand{\arraystretch}{1.15}
\begin{tabular}{cccc}
\hline
\textbf{Method} & \textbf{2M}\,$\uparrow$ & \textbf{3M}\,$\uparrow$ & \textbf{4M}\,$\uparrow$\\
\hline
AdvPatch            & 81.08\% & 35.14\% & 27.03\% \\
AdvTshirt           & 81.08\% & 72.97\% &  8.11\% \\
NatPatch            & 64.86\% & 13.51\% &  0.00\% \\
AdvTexture          & \underline{97.37\%} & 81.58\% & 21.05\% \\
T‑SEA               & 83.78\% & 93.75\% & 81.08\% \\
AdvReal‑YOLOv2      & 81.08\% & 92.86\% & \textbf{97.30\%} \\
AdvReal‑YOLOv3      & 86.49\% & \underline{97.30\%} & \underline{94.59\%} \\
AdvReal‑YOLOv5      & 83.78\% & 94.59\% & \textbf{97.30\%} \\
AdvReal‑FRCNN       & \textbf{100.00\%} & \textbf{100.00\%} & 91.89\% \\
AdvReal‑DDETR       & 86.49\% & 91.89\% & 86.49\% \\
\hline
\end{tabular}}
\label{adequate}
\vspace{-4mm}
\end{table}

\begin{table}[t]
\centering
\caption{Physical‑world ASRs of various methods at different distances under poor lighting conditions}
\resizebox{7cm}{!}{%
\setlength{\tabcolsep}{4pt}
\renewcommand{\arraystretch}{1.15}
\begin{tabular}{cccc}
\hline
\textbf{Method} & \textbf{2M}\,$\uparrow$ & \textbf{3M}\,$\uparrow$ & \textbf{4M}\,$\uparrow$\\
\hline
AdvPatch          & 35.14\% & 12.50\% & 11.11\% \\
AdvTshirt         & \textbf{100.00\%} & 27.78\% & 13.89\% \\
NatPatch          & 2.70\%  & 0.00\%  & 0.00\%  \\
AdvTexture        & \underline{96.67\%} & \underline{83.33\%} & 23.33\% \\
T‑SEA             & 90.00\% & 94.87\% & \underline{97.78\%} \\
AdvReal‑YOLOv2    & 94.12\% & 94.44\% & 34.21\% \\
AdvReal‑YOLOv3    & 52.63\% & 76.32\% & 91.89\% \\
AdvReal‑YOLOv5    & \textbf{100.00\%} & \textbf{100.00\%} & 90.91\% \\
AdvReal‑FRCNN     & 94.59\% & \textbf{100.00\%} & \textbf{100.00\%} \\
AdvReal‑DDETR     & 53.33\% & 86.67\% & 93.33\% \\
\hline
\end{tabular}}
\label{poor}
\vspace{-4mm}
\end{table}

\subsubsection{\textbf{Different distances attacks}}
Tab.~\ref{adequate} and Tab.~\ref{poor} evaluate the attack performance of clothing printed with different patches under different lighting conditions (adequate light and poor light) and different distances (2m, 3m, 4m). Each patch was taken 111 times at different distances, and the victim model was YOLOv5.
In experiments with different distances, the ASR of traditional methods (such as AdvPatch and NatPatch) showed a significant downward trend with increasing distance. However, AdvReal remained stable or even increased. This shows that the distance has little effect on the attack effect. In the case of poor lighting, the method in this paper will have an increase in the attack success rate with increasing distance. This shows that our patch does not depend on certain specific pixels and has very low resolution requirements. It has an attack effect on a macro scale. In the lighting experiment, the ASR of traditional methods dropped sharply in poor light conditions compared to adequate light. However, the attack effects of TSEA, AdvTexture and our method remained stable. This shows that lighting can cause the failure of target detectors and patches. However, the experimental results of the traditional method show that the failure of the patch is dominant. Therefore, the experimental results show that our patch has excellent physical robustness.For the adversarial patches of this paper trained with different target detectors. DDETR is used as a target detector of the transfromer architecture. The attack effect is worse than other target detectors based on convolutional neural networks. It is more robust under attack.

The experimental results show that even under the combined interference of dynamic changes in detection distance and sudden changes in light intensity. The adversarial clothing generation method proposed in this paper still has strong robustness and excellent attack performance.

\begin{figure*}[hb!]
\centering
\includegraphics[width=16 cm]{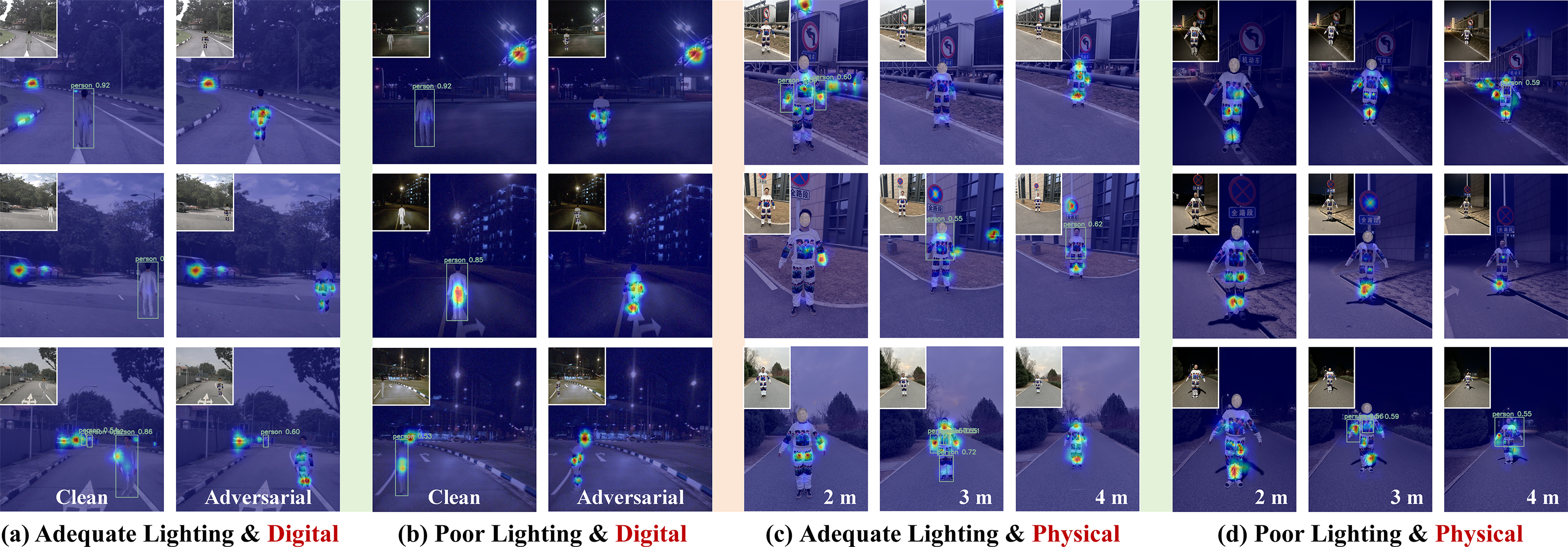}
\caption{Visual evaluation using Grad-CAM in the digital and physical worlds. (a) and (b) are visualization results under different lighting conditions in the digital world. (c) and (d) are visualization results at different distances in the physical world.}
\label{phy_cloud}
\vspace{-4 mm} 
\end{figure*}



\begin{figure}[ht!]
\centering
\includegraphics[width=10 cm]{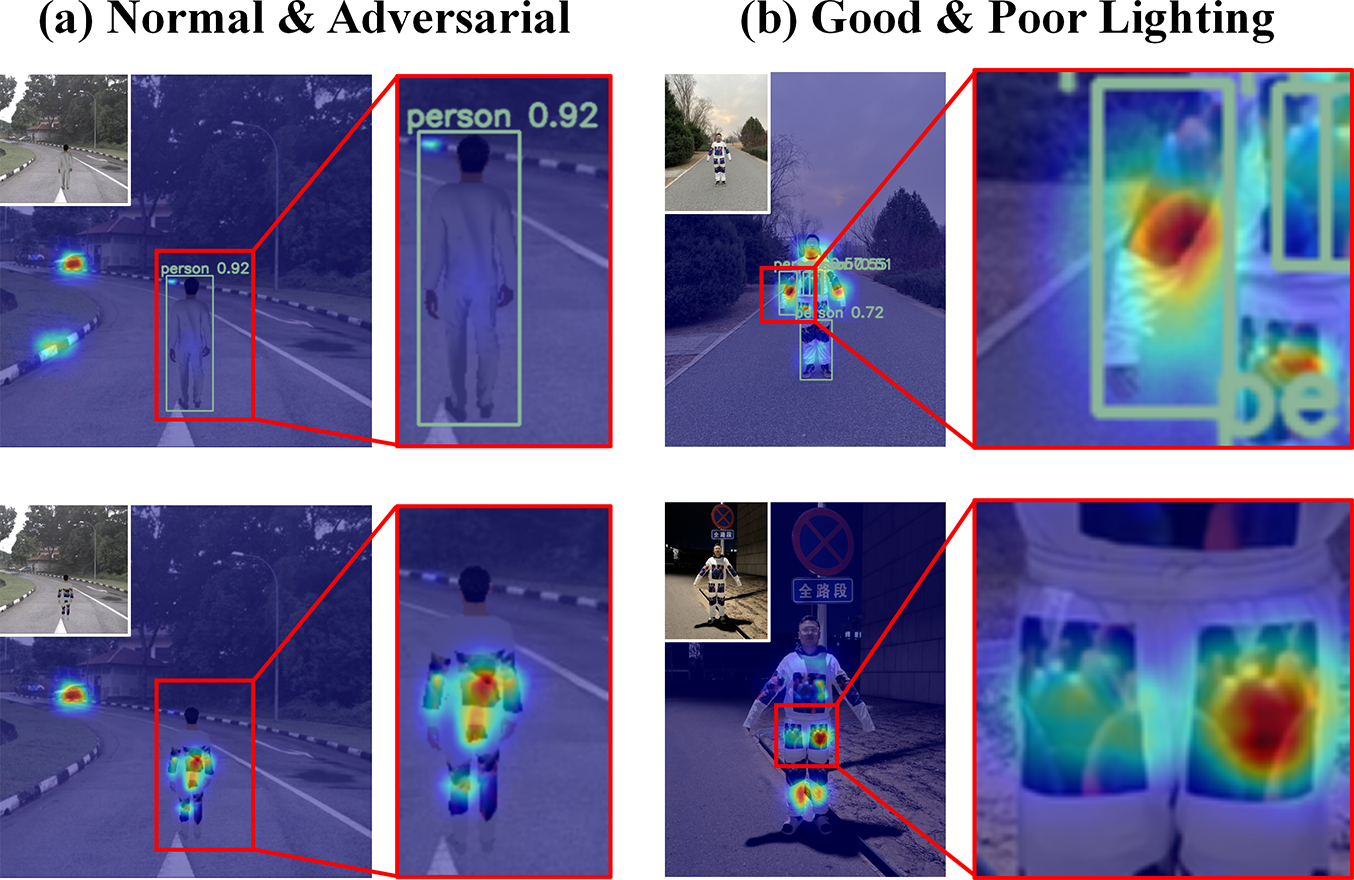}
\caption{Localized attention shift visualized by Grad-CAM with zoom-in regions.}
\label{heatmap_feature}
\vspace{-3 mm} 
\end{figure}

\begin{figure}[ht!]
\centering
\includegraphics[width=10 cm]{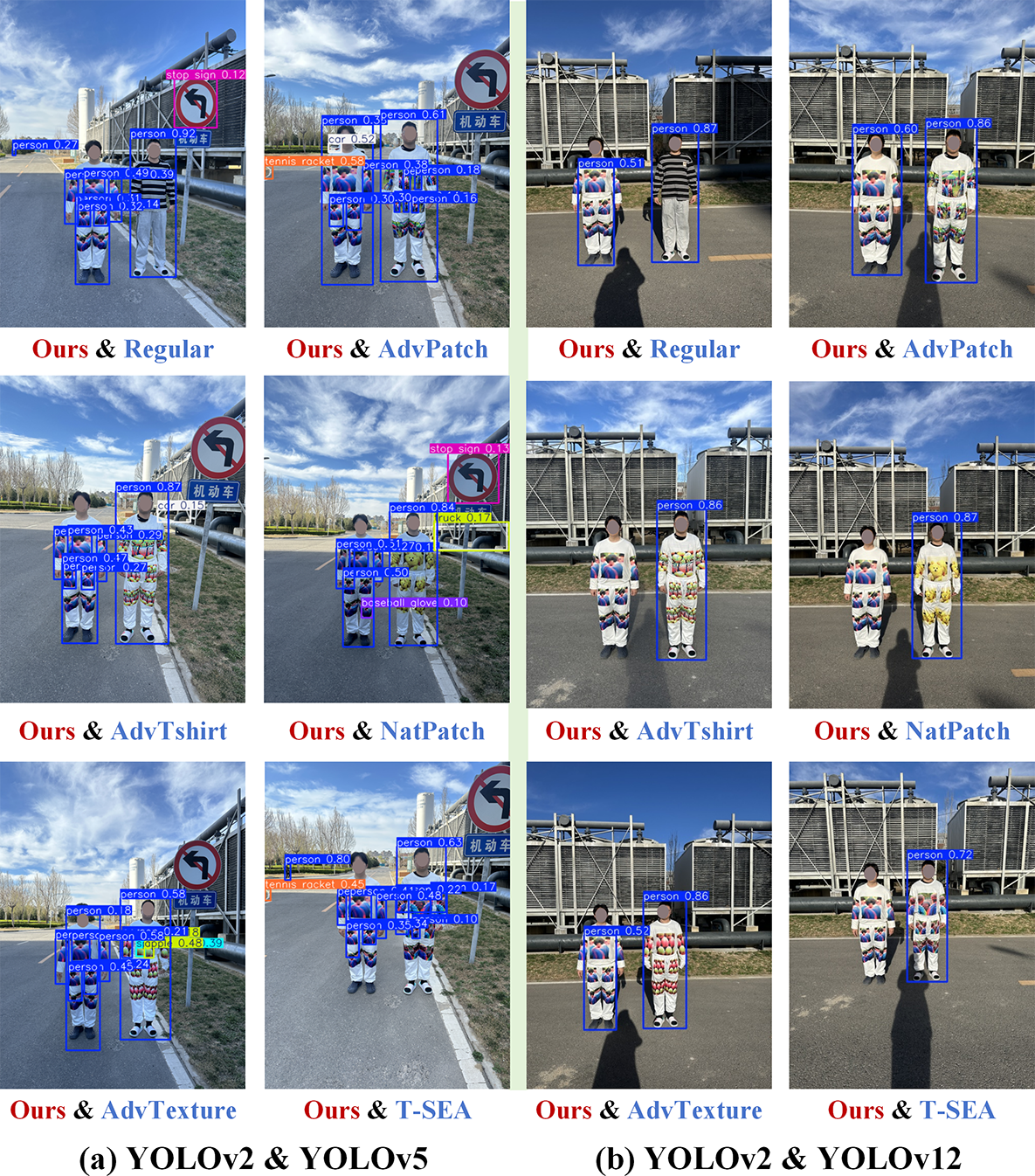}
\caption{Visualization of the proposed adversarial patch and other adversarial patches in the physical world. (a) Glass-box is YOLOv2, closed-box is YOLOv5, and the confidence and IOU thresholds are set to 0.1. (b) Glass-box is YOLOv2, closed-box is YOLOv12, and the confidence and IOU thresholds are set to 0.5.}
\label{phy_vis}
\vspace{-3 mm} 
\end{figure}


\subsubsection{\textbf{Visualization and Interpretability}}
\label{visualization}

We employed Grad-CAM~\cite{selvaraju2017grad} to generate class-discriminative activation maps that reveal the regions the detector attends to when performing person detection. The visualizations provide insight into how adversarial patches disrupt the detector’s semantic attention mechanisms, thereby influencing its decision-making process.

Fig.~\ref{phy_cloud}(a) and (b) present visualizations under good and poor lighting conditions in the digital domain. In the non-adversarial image, the detector’s attention is broadly distributed across the pedestrian's torso and head, aligning with meaningful semantic regions. However, after introducing the adversarial patch, the attention distinctly shifts to the patch region. Fig.~\ref{heatmap_feature} (a) demonstrates that the adversarial patch significantly redirects the detector’s focus from the full-body region to patch-dominated areas. Fig.~\ref{heatmap_feature} (b) shows that despite changes in lighting, the patch consistently captures strong attention responses. Red boxes indicate zoomed-in views of the attended regions. Even under poor illumination, the patch continues to dominate the heatmap, suggesting robustness against lighting variation.

In the physical-world results shown in Fig.~\ref{phy_cloud}(c) and (d), similar behavior is observed at different distances. The detector’s attention weakens around the pedestrian body and intensifies around the patch location, especially when the patch is closer to the camera. This attention hijacking effect illustrates how AdvReal maintains adversarial influence under realistic constraints.

Fig.~\ref{phy_vis}(a) and (b) compare detection responses across detectors under varying confidence and IoU thresholds. As the detector architecture becomes more advanced, its reliance on semantically aligned features increases, making attention shifts caused by the patch more semantically disruptive. Despite this, AdvReal maintains its effectiveness by consistently redirecting attention in all cases.

We also observe that the generated patches often contain structured gradients, edge-like formations, or repetitive textures. These features may partially resemble elements of the human form (e.g., limbs, contours), allowing the patch to falsely activate neuron clusters that normally respond to human-like regions. This suggests that AdvReal does not merely add noise, but introduces targeted feature-level deception.

The observations not only confirm the effectiveness of the attack but also offer interpretability of how the patch operates at the attention level. Such understanding could potentially inform future defense strategies—e.g., attention consistency regularization or saliency-guided anomaly suppression—to mitigate patch-based adversarial threats.

\section{Discussion}

\subsection{Advantage of Proposed Adversarial Algorithm}

\subsubsection{\textbf{Comparative of proposed and other algorithms}}

Existing physical adversarial methods often overlook the significant discrepancies between digital simulations and real-world scenarios, as well as the inherent diversity and uncertainty present in real environments. As a result, their performance tends to be unstable in practical applications. To address the limitations, we proposed \textit{AdvReal}, which offers the following key advantages: 1) \textbf{Joint 2D-3D Optimization:} AdvReal introduces the first joint optimization framework that integrates both 2D and 3D spaces, along with the Shakedrop mechanism, substantially enhancing the transferability of adversarial patches from digital to physical environments. 2) \textbf{Realism-Aware Modeling:} The framework explicitly accounts for realistic conditions, such as illumination changes and human body scale variations, ensuring robust and stable attack performance during real-world testing. 3) \textbf{Multi-View Robustness:} In contrast to methods trained solely on 2D datasets, AdvReal supports multi-view adversarial optimization, enabling it to maintain strong attack effectiveness across a wide range of viewing angles and geometric distortions.

\subsubsection{\textbf{Why 2D–3D joint optimization is effective?}}

single 2D adversarial optimization often relies on static, idealized patches, while real-world scenarios involve depth, occlusion, perspective distortion, and motion—elements better captured by 3D modeling. Optimizing in both 2D\&3D spaces ensures that adversarial patterns remain effective when projected onto real-world surfaces (e.g., curved clothing, variable human angle). 

In physical settings, lighting, scaling, and non-rigi0d modeling play a huge role. 3D rendering allows simulation of these transformations during training, making the patch more robust. Specifically, Tab.~\ref{robustness} shows that AdvReal maintains strong adversarial effectiveness even when large portions of the adversarial patches are occluded. Additionally, Tab.~\ref{adequate} and Tab.~\ref{poor} confirm that AdvReal consistently achieves strong adversarial performance even under challenging illumination conditions. Real-world detectors see targets from all angles, and joint optimization helps enforce effectiveness from multiple viewpoints. Fig.~\ref{angle} further illustrates performance across various viewing angles, highlighting that AdvReal achieves excellent attack results on YOLO detectors, despite slightly reduced effectiveness on robust detectors such as Faster R-CNN and D-DETR. In summary, 3D optimization can effectively supplement the shortcomings of traditional adversarial optimization methods in the physical world.

\subsubsection{\textbf{Discoveries and recommendations of AI-security in object detection system for AVs}}

The results reveal that the vulnerability of object detectors to physical adversarial patches varies considerably across different architectures. Single-stage detectors such as YOLOv11 and YOLOv12 are more susceptible to adversarial manipulation, while the earlier YOLOv8 version demonstrates higher robustness (Tab.~\ref{compare} and Fig.~\ref{confusion}). Two-stage detectors like Faster R-CNN exhibit greater resistance to physical attacks, likely due to their larger number of parameters and the presence of an intermediate region proposal mechanism that helps filter out adversarial cues~\cite{ren2016fasterrcnnrealtimeobject}. Meanwhile, transformer-based models such as D-DETR show even stronger robustness under multi-view conditions, maintaining stable performance across varying viewpoints (Fig.~\ref{angle}).

The differences in adversarial vulnerability can be explained by the architectural nature of the detectors. One-stage models like YOLOv2 and YOLOv3 lack intermediate proposal filtering mechanisms and generate dense predictions directly from feature maps, making them especially sensitive to localized perturbations such as adversarial patches~\cite{im2022adversarial}. In contrast, two-stage detectors (e.g., Faster R-CNN) use a region proposal network to first identify high-confidence candidate regions, which acts as a form of internal filtering that helps suppress adversarial noise~\cite{wang2020adversarial}. Transformer-based detectors (e.g., D-DETR) employ global self-attention mechanisms that further dilute the effect of spatially localized patches~\cite{shao2021adversarial}. These structural differences account for the ASR variations reported in Tab.~\ref{compare} and offer valuable insight for designing security-aware detectors in AVs.

In the context of autonomous driving, image perception systems must strike a balance between real-time processing and resilience to adversarial threats. The experiments indicate that adversarial patches can significantly redirect the detector's attention away from semantically meaningful regions, resulting in missed detections (Fig.~\ref{phy_cloud} and Fig.~\ref{phy_vis}). To mitigate these risks, we recommend reinforcing the structural robustness of detection models. Architectures with multi-stage processing and transformer-based designs demonstrate stronger spatial contextual modeling and higher resistance to adversarial interference. Moreover, detection heads should be equipped with mechanisms such as attention stabilization modules or feature-level denoising blocks to counteract patch-induced interference. In summary, the development of security-aware detectors requires not only stronger model architectures but also defense strategies informed by empirical adversarial behavior in realistic environments.

\subsubsection{\textbf{Toward Identifying Malicious Attacks in AV Perception}}

Although adversarial attacks and natural detection errors may both lead to failed recognition, their underlying characteristics differ in systematic ways that can support intent-level distinction. Based on the and detectability in Sec.~\ref{sec_detectability} and visualizations experiments in Sec.~\ref{visualization} , we observe that adversarial patches consistently redirect model attention to semantically irrelevant regions, whereas natural false positives usually result from ambiguous object boundaries or poor illumination without attention convergence. Furthermore, detection-based defense tools such as NAPGuard show higher response activations on structurally aggressive patterns, which are typically absent in naturally missed detections. These findings suggest that malicious attacks often exhibit spatial consistency, saliency concentration, and abnormal high resolution texture distribution. Thus, distinguishing adversarial attacks from natural false is feasible through joint analysis of attention behavior, patch localization, and detectability signals.

\subsection{Limitations of Proposed Adversarial Algorithm}

Although AdvReal demonstrates superior adversarial performance compared to other methods in both digital and physical experiments, it still has certain potential limitations.

\textbf{(a) Limitations of the attack mechanism.} The attack success rate 0of the adversarial patch generated by the glass-box attack in a closed-box detector is significantly reduced, especially when targeting two-stage and transformer-based detectors. In these cases, the attack success rate of the single-stage adversarial patch drops to a minimum of $37.45 \%$ (Fig.~\ref{confusion}). This is a common problem faced by current adversarial attacks. The difference in feature space between the unified feature map prediction of the single-stage detector and the step-by-step feature processing in the two-stage detector prevents effective alignment of the adversarial perturbation information. \textit{Shakedrop} and other strategy have slightly improved the transferability by changing the complexity of the model. However, a generalized training framework is still required to address the limitation of over-reliance on adversarial attacks.

\textbf{(b) Fight against the shape constraints of the patch.} The proposed AdvReal method mainly focuses on adversarial effectiveness and does not effectively constrain the nature looking of generated patches. Consequently, adversarial patches may appear visually unnatural and easily detectable, limiting their use in scenarios requiring visual subtlety. Future research could further explore achieving a better balance between adversarial effectiveness and nature looking.

\section{Conclusion and Future Works}

This study proposed AdvReal, a robust adversarial patch generation framework designed to address challenges in physical-world attacks on object detection systems, including intra-class diversity, multi-angle variations, and complex environmental conditions. By incorporating 2D–3D joint optimization, physics-driven non-rigid surfaces modeling, and a realistic matching mechanism for lighting and geometry, AdvReal effectively bridges the gap between digital simulations and physical scenarios, thereby enhancing the transferability and robustness of adversarial patches. Experimental results across diverse object detectors and real-world settings demonstrate that AdvReal achieves state-of-the-art attack success rates, superior resilience to occlusion and lighting variations, and notable efficacy in redirecting detector attention. Beyond attack performance, this study also incorporates adversarial detectability evaluation as a potential defense perspective and discusses architectural strategies such as attention stabilization to improve model robustness. The proposed framework underscores the critical role of realistic 3D modeling and joint training in overcoming the limitations of conventional methods, providing a foundational benchmark for evaluating the adversarial robustness of AVs perception systems.

Looking ahead, several promising research directions warrant exploration to further enhance the framework and address current limitations. First, improving the generalization capability of adversarial patches across heterogeneous detector architectures requires consideration of structural differences in feature extraction and decision-making processes. Second, balancing adversarial effectiveness with visual naturalness could lead to more covert and practical attacks. Lastly, insights derived from AdvReal may inform the design of robust defense mechanisms. In particular, integrating adversarial training with patch-aware detection modules or developing co-evolution frameworks for joint attack-defense optimization offers promising potential for enhancing the security of real-world AV perception systems.

\section{Acknowledgments}

This work was supported by the fundamental research funds for the central universities.

\vspace{-4 mm}

\bibliographystyle{elsarticle-harv} 
\bibliography{main}

\begin{thebibliography}{10}
\providecommand{\url}[1]{#1}
\csname url@samestyle\endcsname
\providecommand{\newblock}{\relax}
\providecommand{\bibinfo}[2]{#2}
\providecommand{\BIBentrySTDinterwordspacing}{\spaceskip=0pt\relax}
\providecommand{\BIBentryALTinterwordstretchfactor}{4}
\providecommand{\BIBentryALTinterwordspacing}{\spaceskip=\fontdimen2\font plus
\BIBentryALTinterwordstretchfactor\fontdimen3\font minus \fontdimen4\font\relax}
\providecommand{\BIBforeignlanguage}[2]{{%
\expandafter\ifx\csname l@#1\endcsname\relax
\typeout{** WARNING: IEEEtran.bst: No hyphenation pattern has been}%
\typeout{** loaded for the language `#1'. Using the pattern for}%
\typeout{** the default language instead.}%
\else
\language=\csname l@#1\endcsname
\fi
#2}}
\providecommand{\BIBdecl}{\relax}
\BIBdecl

\bibitem{brown2017adversarial}
T.~B. Brown, D.~Man{\'e}, A.~Roy, M.~Abadi, and J.~Gilmer, ``Adversarial patch,'' \emph{arXiv preprint arXiv:1712.09665}, 2017.

\bibitem{williams2024camopatch}
P.~Williams and K.~Li, ``Camopatch: An evolutionary strategy for generating camoflauged adversarial patches,'' \emph{Advances in Neural Information Processing Systems}, vol.~36, 2024.

\bibitem{zhou2024stealthy}
M.~Zhou, W.~Zhou, J.~Huang, J.~Yang, M.~Du, and Q.~Li, ``Stealthy and effective physical adversarial attacks in autonomous driving,'' \emph{IEEE Transactions on Information Forensics and Security}, 2024.

\bibitem{wei2022simultaneously}
X.~Wei, Y.~Guo, J.~Yu, and B.~Zhang, ``Simultaneously optimizing perturbations and positions for black-box adversarial patch attacks,'' \emph{IEEE transactions on pattern analysis and machine intelligence}, vol.~45, no.~7, pp. 9041--9054, 2022.

\bibitem{liu2024eap}
X.~Liu, F.~Shen, J.~Zhao, and C.~Nie, ``Eap: An effective black-box impersonation adversarial patch attack method on face recognition in the physical world,'' \emph{Neurocomputing}, vol. 580, p. 127517, 2024.

\bibitem{zhou2025fooling}
D.~Zhou, H.~Qu, N.~Wang, C.~Peng, Z.~Ma, X.~Yang, and X.~Gao, ``Fooling human detectors via robust and visually natural adversarial patches,'' \emph{Neurocomputing}, vol. 616, p. 128915, 2025.

\bibitem{abed2024deep}
A.~Abed, B.~Akrout, and I.~Amous, ``Deep learning-based few-shot person re-identification from top-view rgb and depth images,'' \emph{Neural Computing and Applications}, vol.~36, no.~31, pp. 19\,365--19\,382, 2024.

\bibitem{thys2019fooling}
S.~Thys, W.~Van~Ranst, and T.~Goedem{\'e}, ``Fooling automated surveillance cameras: adversarial patches to attack person detection,'' in \emph{Proceedings of the IEEE/CVF conference on computer vision and pattern recognition workshops}, 2019, pp. 0--0.

\bibitem{ren2016faster}
S.~Ren, K.~He, R.~Girshick, and J.~Sun, ``Faster r-cnn: Towards real-time object detection with region proposal networks,'' \emph{IEEE transactions on pattern analysis and machine intelligence}, vol.~39, no.~6, pp. 1137--1149, 2016.

\bibitem{redmon2018yolov3}
J.~Redmon, ``Yolov3: An incremental improvement,'' \emph{arXiv preprint arXiv:1804.02767}, 2018.

\bibitem{carion2020end}
N.~Carion, F.~Massa, G.~Synnaeve, N.~Usunier, A.~Kirillov, and S.~Zagoruyko, ``End-to-end object detection with transformers,'' in \emph{European conference on computer vision}.\hskip 1em plus 0.5em minus 0.4em\relax Springer, 2020, pp. 213--229.

\bibitem{hu2022adversarial}
Z.~Hu, S.~Huang, X.~Zhu, F.~Sun, B.~Zhang, and X.~Hu, ``Adversarial texture for fooling person detectors in the physical world,'' in \emph{Proceedings of the IEEE/CVF conference on computer vision and pattern recognition}, 2022, pp. 13\,307--13\,316.

\bibitem{maesumi2021learning}
A.~Maesumi, M.~Zhu, Y.~Wang, T.~Chen, Z.~Wang, and C.~Bajaj, ``Learning transferable 3d adversarial cloaks for deep trained detectors,'' \emph{arXiv preprint arXiv:2104.11101}, 2021.

\bibitem{xu2020adversarial}
K.~Xu, G.~Zhang, S.~Liu, Q.~Fan, M.~Sun, H.~Chen, P.-Y. Chen, Y.~Wang, and X.~Lin, ``Adversarial t-shirt! evading person detectors in a physical world,'' in \emph{Computer Vision--ECCV 2020: 16th European Conference, Glasgow, UK, August 23--28, 2020, Proceedings, Part V 16}.\hskip 1em plus 0.5em minus 0.4em\relax Springer, 2020, pp. 665--681.

\bibitem{chow2020adversarial}
K.-H. Chow, L.~Liu, M.~Loper, J.~Bae, M.~E. Gursoy, S.~Truex, W.~Wei, and Y.~Wu, ``Adversarial objectness gradient attacks in real-time object detection systems,'' in \emph{2020 Second IEEE International Conference on Trust, Privacy and Security in Intelligent Systems and Applications (TPS-ISA)}.\hskip 1em plus 0.5em minus 0.4em\relax IEEE, 2020, pp. 263--272.

\bibitem{hu2021naturalistic}
Y.-C.-T. Hu, B.-H. Kung, D.~S. Tan, J.-C. Chen, K.-L. Hua, and W.-H. Cheng, ``Naturalistic physical adversarial patch for object detectors,'' in \emph{Proceedings of the IEEE/CVF International Conference on Computer Vision}, 2021, pp. 7848--7857.

\bibitem{guesmi2024dap}
A.~Guesmi, R.~Ding, M.~A. Hanif, I.~Alouani, and M.~Shafique, ``Dap: A dynamic adversarial patch for evading person detectors,'' in \emph{Proceedings of the IEEE/CVF Conference on Computer Vision and Pattern Recognition}, 2024, pp. 24\,595--24\,604.

\bibitem{jing2024pad}
L.~Jing, R.~Wang, W.~Ren, X.~Dong, and C.~Zou, ``Pad: Patch-agnostic defense against adversarial patch attacks,'' in \emph{Proceedings of the IEEE/CVF Conference on Computer Vision and Pattern Recognition}, 2024, pp. 24\,472--24\,481.

\bibitem{hu2023physically}
Z.~Hu, W.~Chu, X.~Zhu, H.~Zhang, B.~Zhang, and X.~Hu, ``Physically realizable natural-looking clothing textures evade person detectors via 3d modeling,'' in \emph{Proceedings of the IEEE/CVF Conference on Computer Vision and Pattern Recognition}, 2023, pp. 16\,975--16\,984.

\bibitem{hingun2023reap}
N.~Hingun, C.~Sitawarin, J.~Li, and D.~Wagner, ``Reap: a large-scale realistic adversarial patch benchmark,'' in \emph{Proceedings of the IEEE/CVF International Conference on Computer Vision}, 2023, pp. 4640--4651.

\bibitem{caesar2020nuscenes}
H.~Caesar, V.~Bankiti, A.~H. Lang, S.~Vora, V.~E. Liong, Q.~Xu, A.~Krishnan, Y.~Pan, G.~Baldan, and O.~Beijbom, ``nuscenes: A multimodal dataset for autonomous driving,'' in \emph{Proceedings of the IEEE/CVF conference on computer vision and pattern recognition}, 2020, pp. 11\,621--11\,631.

\bibitem{redmon2017yolo9000}
J.~Redmon and A.~Farhadi, ``Yolo9000: better, faster, stronger,'' in \emph{Proceedings of the IEEE conference on computer vision and pattern recognition}, 2017, pp. 7263--7271.

\bibitem{redmon2018yolov3incrementalimprovement}
\BIBentryALTinterwordspacing
------, ``Yolov3: An incremental improvement,'' 2018. [Online]. Available: \url{https://arxiv.org/abs/1804.02767}
\BIBentrySTDinterwordspacing

\bibitem{yolov5}
G.~Jocher, A.~Stoken, J.~Borovec, L.~Changyu, A.~Hogan, L.~Diaconu, J.~Poznanski, L.~Yu, P.~Rai, R.~Ferriday \emph{et~al.}, ``ultralytics/yolov5: v3. 0,'' \emph{Zenodo}, 2020.

\bibitem{ddetr}
X.~Zhu, W.~Su, L.~Lu, B.~Li, X.~Wang, and J.~Dai, ``Deformable detr: Deformable transformers for end-to-end object detection,'' \emph{arXiv preprint arXiv:2010.04159}, 2020.

\bibitem{ren2016fasterrcnnrealtimeobject}
\BIBentryALTinterwordspacing
S.~Ren, K.~He, R.~Girshick, and J.~Sun, ``Faster r-cnn: Towards real-time object detection with region proposal networks,'' 2016. [Online]. Available: \url{https://arxiv.org/abs/1506.01497}
\BIBentrySTDinterwordspacing

\bibitem{huang2023t}
H.~Huang, Z.~Chen, H.~Chen, Y.~Wang, and K.~Zhang, ``T-sea: Transfer-based self-ensemble attack on object detection,'' in \emph{Proceedings of the IEEE/CVF conference on computer vision and pattern recognition}, 2023, pp. 20\,514--20\,523.

\bibitem{selvaraju2017grad}
R.~R. Selvaraju, M.~Cogswell, A.~Das, R.~Vedantam, D.~Parikh, and D.~Batra, ``Grad-cam: Visual explanations from deep networks via gradient-based localization,'' in \emph{Proceedings of the IEEE international conference on computer vision}, 2017, pp. 618--626.

\bibitem{wang2022fca}
D.~Wang, T.~Jiang, J.~Sun, W.~Zhou, Z.~Gong, X.~Zhang, W.~Yao, and X.~Chen, ``Fca: Learning a 3d full-coverage vehicle camouflage for multi-view physical adversarial attack,'' in \emph{Proceedings of the AAAI conference on artificial intelligence}, vol.~36, no.~2, 2022, pp. 2414--2422.

\bibitem{zhang2022intelligent}
X.~Zhang, J.~Huang, Y.~Huang, K.~Huang, L.~Yang, Y.~Han, L.~Wang, H.~Liu, J.~Luo, and J.~Li, ``Intelligent amphibious ground-aerial vehicles: State of the art technology for future transportation,'' \emph{IEEE Transactions on Intelligent Vehicles}, vol.~8, no.~1, pp. 970--987, 2022.

\bibitem{bai2024ar}
X.~Bai, P.~Dong, Y.~Huang, S.~Kumari, H.~Yu, and Y.~Ren, ``An ar-based meta vehicle road cooperation testing systems: framework, components modeling and an implementation example,'' \emph{IEEE Internet of Things Journal}, 2024.

\bibitem{swerdlow2024street}
A.~Swerdlow, R.~Xu, and B.~Zhou, ``Street-view image generation from a bird's-eye view layout,'' \emph{IEEE Robotics and Automation Letters}, 2024.

\bibitem{huang2025advswap}
Y.~Huang, Q.~Zhang, J.~Xing, M.~Cheng, H.~Yu, Y.~Ren, and X.~Xiong, ``Advswap: Covert adversarial perturbation with high frequency info-swapping for autonomous driving perception,'' in \emph{2024 IEEE 27th International Conference on Intelligent Transportation Systems (ITSC)}, 2024, pp. 1686--1693.

\bibitem{wang2024attention}
L.~Wang, J.~Huang, L.~Huang, F.~Wang, C.~Gao, J.~Li, F.~Xiao, and D.~Luo, ``Attention-disentangled re-id network for unsupervised domain adaptive person re-identification,'' \emph{Knowledge-Based Systems}, vol. 304, p. 112583, 2024.

\bibitem{cui2024adversarial}
J.~Cui, W.~Guo, H.~Huang, X.~Lv, H.~Cao, and H.~Li, ``Adversarial examples for vehicle detection with projection transformation,'' \emph{IEEE Transactions on Geoscience and Remote Sensing}, 2024.

\bibitem{mahima2024toward}
K.~Y. Mahima, A.~G. Perera, S.~Anavatti, and M.~Garratt, ``Toward robust 3d perception for autonomous vehicles: A review of adversarial attacks and countermeasures,'' \emph{IEEE Transactions on Intelligent Transportation Systems}, 2024.

\bibitem{huang2016deep}
G.~Huang, Y.~Sun, Z.~Liu, D.~Sedra, and K.~Q. Weinberger, ``Deep networks with stochastic depth,'' in \emph{Computer Vision--ECCV 2016: 14th European Conference, Amsterdam, The Netherlands, October 11--14, 2016, Proceedings, Part IV 14}.\hskip 1em plus 0.5em minus 0.4em\relax Springer, 2016, pp. 646--661.

\bibitem{li2025uv}
Y.~Li, W.~Zhang, K.~Liang, and B.~Xiao, ``Uv-attack: Physical-world adversarial attacks for person detection via dynamic-nerf-based uv mapping,'' \emph{arXiv preprint arXiv:2501.05783}, 2025.

\bibitem{wei2024physical}
H.~Wei, H.~Tang, X.~Jia, Z.~Wang, H.~Yu, Z.~Li, S.~Satoh, L.~Van~Gool, and Z.~Wang, ``Physical adversarial attack meets computer vision: A decade survey,'' \emph{IEEE Transactions on Pattern Analysis and Machine Intelligence}, 2024.

\bibitem{agudo2017force}
A.~Agudo and F.~Moreno-Noguer, ``Force-based representation for non-rigid shape and elastic model estimation,'' \emph{IEEE transactions on pattern analysis and machine intelligence}, vol.~40, no.~9, pp. 2137--2150, 2017.

\bibitem{gundogdu2020garnet}
E.~Gundogdu, V.~Constantin, S.~Parashar, A.~Seifoddini, M.~Dang, M.~Salzmann, and P.~Fua, ``Garnet++: Improving fast and accurate static 3d cloth draping by curvature loss,'' \emph{IEEE Transactions on Pattern Analysis and Machine Intelligence}, vol.~44, no.~1, pp. 181--195, 2020.

\bibitem{yolov8_ultralytics}
\BIBentryALTinterwordspacing
G.~Jocher, A.~Chaurasia, and J.~Qiu, ``Ultralytics yolov8,'' 2023. [Online]. Available: \url{https://github.com/ultralytics/ultralytics}
\BIBentrySTDinterwordspacing

\bibitem{yolo11_ultralytics}
\BIBentryALTinterwordspacing
G.~Jocher and J.~Qiu, ``Ultralytics yolo11,'' 2024. [Online]. Available: \url{https://github.com/ultralytics/ultralytics}
\BIBentrySTDinterwordspacing

\bibitem{tian2025yolov12}
Y.~Tian, Q.~Ye, and D.~Doermann, ``Yolov12: Attention-centric real-time object detectors,'' \emph{arXiv preprint arXiv:2502.12524}, 2025.

\bibitem{yamada2018shakedrop}
\BIBentryALTinterwordspacing
Y.~Yamada, M.~Iwamura, and K.~Kise, ``Shakedrop regularization,'' 2018. [Online]. Available: \url{https://openreview.net/forum?id=S1NHaMW0b}
\BIBentrySTDinterwordspacing

\bibitem{dalal2005histograms}
N.~Dalal and B.~Triggs, ``Histograms of oriented gradients for human detection,'' in \emph{2005 IEEE computer society conference on computer vision and pattern recognition (CVPR'05)}, vol.~1.\hskip 1em plus 0.5em minus 0.4em\relax Ieee, 2005, pp. 886--893.

\bibitem{lin2014microsoft}
T.-Y. Lin, M.~Maire, S.~Belongie, J.~Hays, P.~Perona, D.~Ramanan, P.~Doll{\'a}r, and C.~L. Zitnick, ``Microsoft coco: Common objects in context,'' in \emph{Computer vision--ECCV 2014: 13th European conference, zurich, Switzerland, September 6-12, 2014, proceedings, part v 13}.\hskip 1em plus 0.5em minus 0.4em\relax Springer, 2014, pp. 740--755.

\bibitem{ran2025black}
Y.~Ran, A.-X. Zhang, M.~Li, W.~Tang, and Y.-G. Wang, ``Black-box adversarial attacks against image quality assessment models,'' \emph{Expert Systems with Applications}, vol. 260, p. 125415, 2025.

\bibitem{giri2025so}
K.~J. Giri \emph{et~al.}, ``So-yolov8: A novel deep learning-based approach for small object detection with yolo beyond coco,'' \emph{Expert Systems with Applications}, p. 127447, 2025.

\bibitem{fang2024state}
J.~Fang, Y.~Jiang, C.~Jiang, Z.~L. Jiang, C.~Liu, and S.-M. Yiu, ``State-of-the-art optical-based physical adversarial attacks for deep learning computer vision systems,'' \emph{Expert Systems with Applications}, p. 123761, 2024.

\bibitem{zhang2021evaluating}
J.~Zhang, Y.~Lou, J.~Wang, K.~Wu, K.~Lu, and X.~Jia, ``Evaluating adversarial attacks on driving safety in vision-based autonomous vehicles,'' \emph{IEEE Internet of Things Journal}, vol.~9, no.~5, pp. 3443--3456, 2021.

\bibitem{xiao2025transformer}
C.~Xiao, S.~Peng, L.~Zhang, J.~Wang, D.~Ding, and J.~Zhang, ``A transformer-based adversarial network framework for steganography,'' \emph{Expert Systems with Applications}, vol. 269, p. 126391, 2025.

\bibitem{zhang2021mining}
A.~Zhang, Y.~Liao, S.~Liu, M.~Lu, Y.~Wang, C.~Gao, and X.~Li, ``Mining the benefits of two-stage and one-stage hoi detection,'' \emph{Advances in Neural Information Processing Systems}, vol.~34, pp. 17\,209--17\,220, 2021.

\bibitem{nguyen2025survey}
K.~N.~T. Nguyen, W.~Zhang, K.~Lu, Y.-H. Wu, X.~Zheng, H.~L. Tan, and L.~Zhen, ``A survey and evaluation of adversarial attacks in object detection,'' \emph{IEEE Transactions on Neural Networks and Learning Systems}, 2025.

\bibitem{im2022adversarial}
J.~Im~Choi and Q.~Tian, ``Adversarial attack and defense of yolo detectors in autonomous driving scenarios,'' in \emph{2022 IEEE intelligent vehicles symposium (IV)}.\hskip 1em plus 0.5em minus 0.4em\relax IEEE, 2022, pp. 1011--1017.

\bibitem{wang2020adversarial}
Y.~Wang, K.~Wang, Z.~Zhu, and F.-Y. Wang, ``Adversarial attacks on faster r-cnn object detector,'' \emph{Neurocomputing}, vol. 382, pp. 87--95, 2020.

\bibitem{shao2021adversarial}
R.~Shao, Z.~Shi, J.~Yi, P.-Y. Chen, and C.-J. Hsieh, ``On the adversarial robustness of vision transformers,'' \emph{arXiv preprint arXiv:2103.15670}, 2021.

\bibitem{ying2020patches}
Z.~Ying, H.~Niu, P.~Gupta, D.~Mahajan, D.~Ghadiyaram, and A.~Bovik, ``From patches to pictures (paq-2-piq): Mapping the perceptual space of picture quality,'' in \emph{Proceedings of the IEEE/CVF conference on computer vision and pattern recognition}, 2020, pp. 3575--3585.

\bibitem{wu2024napguard}
S.~Wu, J.~Wang, J.~Zhao, Y.~Wang, and X.~Liu, ``Napguard: Towards detecting naturalistic adversarial patches,'' in \emph{Proceedings of the IEEE/CVF Conference on Computer Vision and Pattern Recognition}, 2024, pp. 24\,367--24\,376.

\bibitem{wang2023adversarial}
J.~Wang, X.~Liu, J.~Hu, D.~Wang, S.~Wu, T.~Jiang, Y.~Guo, A.~Liu, and J.~Zhou, ``Adversarial examples in the physical world: A survey,'' \emph{arXiv preprint arXiv:2311.01473}, 2023.

\end{thebibliography}

\end{document}